\documentclass[11pt,a4paper]{article}
\usepackage{times,latexsym}
\usepackage{url}
\usepackage[T1]{fontenc}

\usepackage[table]{xcolor}% http://ctan.org/pkg/xcolor

\usepackage{microtype}
\usepackage{graphicx}
\usepackage{booktabs} % for professional tables
\usepackage{array, caption, floatrow, tabularx, makecell, booktabs}
\usepackage{comment}

\usepackage{bbold}
\usepackage{layouts}
\usepackage{enumitem,kantlipsum}
\usepackage{pgf}
\usepackage{import}
\usepackage{caption}
\usepackage{subcaption}
\usepackage{subfiles}
\usepackage{scalefnt}
\usepackage{soul}

\definecolor{ar}{HTML}{045275}
\definecolor{de}{HTML}{089099}
\definecolor{en}{HTML}{7CCBA2}
\definecolor{es}{HTML}{FCDE9C}
\definecolor{ru}{HTML}{F0746E}
\definecolor{sw}{HTML}{DC3977}
\definecolor{zh}{HTML}{7C1D6F}

\definecolor{lape_fs}{HTML}{4fb9af}
\definecolor{lape_pruned}{HTML}{045275}
\definecolor{teal_dark}{HTML}{045275}
\definecolor{navy}{HTML}{e47914}
\definecolor{green}{HTML}{7ccba2}
\definecolor{rev}{HTML}{e07102}
\definecolor{teal-1}{HTML}{008080}
\definecolor{teal-2}{HTML}{008057}
\definecolor{navy-1}{HTML}{e49f14}
\definecolor{navy-2}{HTML}{e47914}

\usepackage{svg}

\usepackage{multirow}
\usepackage{placeins}

\usepackage{amsmath}
\usepackage{amssymb}
\usepackage{mathtools}
\usepackage{amsthm}

\usepackage{tabularx}
\usepackage[export]{adjustbox}

\usepackage[acceptedWithA]{tacl2021v1}
\AtBeginDocument{\thispagestyle{plain}}

\makeatletter
\AddToShipoutPicture{%
  \AtTextUpperLeft{%
    \put(0,\LenToUnit{1cm}){%
      \parbox{\textwidth}{\centering\naaclhv
        Accepted for publication in TACL}}}}
\makeatother

\usepackage{xspace,mfirstuc,tabulary}

\newif\iftaclinstructions
\taclinstructionsfalse % AUTHORS: do NOT set this to true
\iftaclinstructions
\renewcommand{\confidential}{}
\renewcommand{\anonsubtext}{(No author info supplied here, for consistency with
TACL-submission anonymization requirements)}
\newcommand{\instr}
\fi

\iftaclpubformat % this "if" is set by the choice of options

\else

\fi

\title{On the Limitations of Language Targeted Pruning: Investigating the Calibration Language Impact in Multilingual LLM Pruning}

\author{
    Simon Kurz\textsuperscript{1,4} \quad
    Jian-Jia Chen\textsuperscript{1,4} \quad
    Lucie Flek\textsuperscript{2,4} \quad
    Zhixue Zhao\textsuperscript{3} \\
    \textsuperscript{1}Department of Computer Science, TU Dortmund University, Germany \\
    \textsuperscript{2}Bonn-Aachen International Center for Information Technology, University of Bonn, Germany \\
    \textsuperscript{3}Computer Science School, University of Sheffield, United Kingdom \\
    \textsuperscript{4}Lamarr Institute for Machine Learning and Artificial Intelligence, Germany \\
    \texttt{simon.kurz@tu-dortmund.de, zhixue.zhao@sheffield.ac.uk}
}

\date{}

\begin{document}
\maketitle
\begin{abstract}
Recent advances in large language model (LLM) pruning have shown state-of-the-art (SotA) compression results in post-training and retraining-free settings while maintaining high predictive performance. However, previous research mainly considered calibrating based on English text, despite the multilingual nature of modern LLMs and their frequent use in non-English languages. This analysis paper conducts an in-depth investigation of the performance and internal representation changes associated with pruning multilingual language models for monolingual applications. We present the first comprehensive empirical study, comparing different calibration languages for pruning multilingual models across diverse languages, tasks, models, and SotA pruning techniques. We further analyze the latent subspaces, pruning masks, and individual neurons within pruned models. Our results reveal that while calibration on the target language effectively retains perplexity and yields high signal-to-noise ratios, it does not consistently improve downstream task performance. Further analysis of internal representations at three different levels highlights broader limitations of current pruning approaches: While they effectively preserve dominant information like language-specific features, this is insufficient to counteract the loss of nuanced, language-agnostic features that are crucial for knowledge retention and reasoning.
\end{abstract}

\section{Introduction}

State-of-the-art language models often rely on over-parameterization with billions of parameters, resulting in significant memory and computational demands~\citep{zhang2017understanding,allen2019convergence,xu2023over}. 
To address this, model compression methods such as quantization and pruning are commonly used~\citep{gholami2022survey,hoefler2021sparsity,kuzmin2023pruning,frantar-gptq,zhu2023survey,frantar-gptq,liu2025vptq2}. 
Established post-training pruning methods for language models, such as SparseGPT~\citep{frantar2023sparsegpt} and Wanda~\citep{sun2024a}, demonstrated competitive performance in a retraining-free setting~\citep{zhu2023survey} , using only a small calibration dataset instead of big training corpora~\citep{zhu2023survey,kuzmin2022fp8,frantar2023sparsegpt,kuzmin2023pruning}. 

However, current research on LLM pruning mostly focuses on English calibration and evaluation, despite the widespread use of multilingual LLMs for non-English tasks~\citep{touvron2023llama,achiam2023gpt,jiang2023mistral}. For example, if we want to prune a multilingual LM for tasks in French, should we use calibration data in French or English? This disconnect between research focus and real-world usage raises critical questions: \textbf{How does the choice of calibration language impact pruning outcomes for multilingual LLMs for monolingual applications? To what extent and why?} 

We provide the first comprehensive empirical study of how the calibration language affects multilingual LLM pruning. 
We find that \textbf{calibrating on text in the target language minimizes pruning errors and best retains language modeling capabilities related to perplexity} (Section~\ref{sec:subsection_perplexity}). \textbf{However, this does not always lead to better downstream task performance} (Section~\ref{sec:downstream_performance}, Section~\ref{sec:open_domain_ansering}). 
Further, calibration in a different language, despite its distinct linguistic features, can yield better results, suggesting that current pruning approaches give more importance to dominant language-specific features while neglecting subtle, language-agnostic features crucial for reasoning tasks.

Through multi-level analysis spanning latent subspaces, weight matrices, and individual neurons, we investigate the mechanisms behind our observations and derive the following findings. 
\begin{itemize}
    \item Calibrating on target languages consistently yields the lowest perplexity (Section~\ref{sec:subsection_perplexity}) but will not guarantee optimal downstream performance (Section~\ref{sec:downstream_performance}). 
    \item The calibration language is unlikely to impact language-agnostic features associated with understanding, reasoning, and knowledge retrieval. In contrast, the preservation of language-specific features, such as language modeling capabilities related to perplexity, depends on the selection of calibration languages (Section~\ref{sec:analysis_subspace}).
    \item Pruning struggles to consistently identify essential weights in the attention output projections, partly responsible for the model's language-agnostic reasoning capabilities (Section~\ref{sec:analysis_pruning_mask}).
    \item The neuron activation frequency of language-specific neurons in the FFN layers is not reliably preserved after pruning (Section~\ref{sec:analysis_neurons}).
\end{itemize}

\section{Background}

\subsection{Model Pruning}

Pruning~\citep{lecun1989optimal} is a widely used compression approach, removing parameters that contribute least to model performance to reduce the computational and memory footprint. 
Unlike sparse training~\citep{yuan2021mest,hoang2023revisiting,zhang2023epitopological} or pruning-aware training~\citep{liu2021sparse}, which require iterative training to achieve sparsity, post-training pruning eliminates redundant weights directly based on their importance scores computed using a small calibration set, without requiring additional training. 
This retraining-free feature makes post-training pruning a more efficient approach for LLMs~\citep{frantar2023sparsegpt,sun2024a}.

Based on pruning granularity, pruning methods fall into two major categories~\citep{zhu-etal-2024-survey-model,zafrir2021pruneallsparsepretrained,guo2025slimllm}.
\textbf{Structured pruning} removes entire model components such as neurons, attention heads, or even layers, enabling hardware efficiency but limiting flexibility.
\textbf{Unstructured pruning} removes individual weights without pattern constraints, offering high performance at the cost of deployment complexity. 
Its interpretability and flexibility make it the basis for many analytical studies and recent methods.

Post-training pruning estimates weight importance by minimizing the local pruning error between original $Y=WX$ and pruned layer outputs $\hat{Y} = (M \odot W)X$ for weights $W \in R^{m \times n}$, calibration activations $X \in R^{n \times b}$ and binary pruning mask $M$.
Because pruning preserves weights and activations that optimize $argmin_M||Y - \hat{Y}||^2_F$ on the calibration set, the choice of calibration language directly affects which weights get retained. 
For instance, calibration in English biases pruning toward English-specific patterns. 
This issue becomes critical in multilingual models, where features vary between languages~\citep{shamrai-2024-language,xie2022discovering,choenni2025mwandaimprovingoneshotpruning}.

There are two major unstructured pruning methods that still serve as strong benchmarks and foundations for extensions. 
\textbf{SparseGPT}~\citep{frantar2023sparsegpt} iteratively prunes low-importance weights, estimated using local second-order derivative information via a diagonal Hessian approximation as: 

\begin{equation}
    S_{i,j}^{\text{SGPT}} = \frac{W_{i,j}^2}{H^{-1}_{j,j}}, \quad \text{with} \quad H = XX^\top.
\end{equation}

Weights are iteratively pruned by lowest pruning score $S_{i,j}^{\text{SGPT}}$, while using Hessian information $H$ to update the remaining weights, compensating induced errors. 

\textbf{Wanda}~\citep{sun2024a} uses a simpler one-shot heuristic based on weight magnitude and activation norms, omitting weight updates and expensive Hessian inversions: 

\begin{equation}
    S_{i,j}^{\text{Wanda}} = |W_{i,j}| \cdot \|X_j\|_2
\end{equation}

Although less accurate than SparseGPT in theory, Wanda is more efficient and easier to implement. 
Both methods are widely adopted and have inspired numerous extensions, such as Wanda++~\citep{yang2025wandapruninglargelanguage}, and M-Wanda~\citep{choenni2025mwandaimprovingoneshotpruning}, which adapt these core ideas to gradient-informed or multilingual settings.

\subsection{Surface-Level Evaluation Metrics}

Pruned LLMs can be evaluated at the output-level and at the level of representational metrics that aim to quantify the internal consistency of the compressed model. 

\textbf{Perplexity (PPL)} remains a fundamental metric in language modeling. 
It measures the next-token-prediction uncertainty of a model over a sequence of tokens, formally defined as: 

\begin{equation}
    \text{PPL} = \exp\left( -\frac{1}{T} \sum_{t=1}^{T} \log P(w_t \mid w_{<t}) \right),
\end{equation}

where $P(w_t \mid w_{<t})$ is the probability of the next token predicted by the model. 
Lower perplexity indicates more confident predictions, typically interpreted as higher fluency or syntactic correctness. 
Despite its limitations in capturing semantic or reasoning quality~\citep{meister-cotterell-2021-language,jaiswal2024compressing}, PPL remains a standard first indicator of language modeling performance and is widely used to assess the effects of pruning and quantization~\citep{yang2025wandapruninglargelanguage,frantar2023sparsegpt}. 
In multilingual settings, PPL is typically reported per language to reflect pruning performance across language-specific features~\citep{zeng-etal-2024-multilingual,shamrai-2024-language}. 
However,~\citet{poelman-lhoneux-2025-roles} suggest that the performance of language modeling and language understanding does not necessarily reflect the downstream task performance.

A more direct measure of pruning effects is the \textbf{pruning error}, which quantifies the local reconstruction error at the activation level. 
It is defined as the L2 distance between layer outputs before and after pruning: 

\begin{equation}
    E = \|WX - (M \odot W)X\|_2^2
\end{equation}

This formulation aligns with the reconstruction error minimization paradigm used by many pruning methods, including SparseGPT and Wanda \cite{frantar2023sparsegpt,sun2024a}. 
While one pruning error measurement alone only captures local, layer-wise deviations, it does not account for cumulative error propagation effects from one layer to the next. 
Therefore, some studies measure global pruning error over a test set by aggregating activation deviations across all layers, weighted by the number of activation feature elements~\citep{shin2024rethinking,li2025gptaqefficientfinetuningfreequantization}.

\textbf{Signal-to-Noise Ratio (SNR)} is a normalized version of the pruning error, eliminating the effect of different layer output magnitudes between layer components~\citep{kuzmin2023pruning}, define as %$SNR = 10 \times log \left( ||Y||_2^2 \div \|Y - \hat{Y}|_2^2 \right)$.

\begin{equation}
    SNR = 10 \cdot log \left( \frac{||WX||_2^2}{\|WX - (M \odot W)X\|_2^2} \right)
\end{equation}

High SNR values indicate that the retained weights preserve the dominant signal, while low values suggest that pruning has disrupted key feature components.

While PPL, pruning error, and SNR do not reliably capture higher-order capabilities such as reasoning or knowledge retrieval, they are theoretically grounded and still correlated to model performance~\citep{frantar2023sparsegpt,kuzmin2023pruning,shin2024rethinking}, making them surface-level indicators for assessing pruning quality.

\section{Related Work}

\subsection{Multilingual Language Models}
Most SotA Language Models (LMs), such as Llama-3~\citep{meta2024introducing}, Phi 3~\citep{abdin2024phi}, and DeepSeek-R1~\citep{deepseekai2025}, are trained on multilingual data, enabling them to process and generate text across multiple languages~\citep{huang-etal-2023-languages,holmstrom-etal-2023-bridging,xu2024survey,meta2024introducing}. 
Although multilingual LMs follow the general training paradigm of monolingual LMs, they often exhibit distinct behaviors and unique characteristics~\citep{xu2024survey}. For example, \citet{deng2024multilingual} reveal that multilingual LMs are prone to generate unsafe outputs on malicious instructions in non-English languages, i.e. multilingual jailbreak. \citet{DBLP:journals/corr/abs-2310-00905} report significantly higher rates of unsafe responses to non-English queries.  
\citet{chen-etal-2024-monolingual} find that instruction tuning in a multilingual setting performs on par with, or even surpasses, tuning a model individually for each language. 
Furthermore, previous work on model explanation finds that multilingual models have different process mechanisms from counterparts monolingual models~\citep{jorgensen-etal-2022-multilingual,zhao-aletras-2024-comparing}. 

In other areas of multilingual LLM research, such as cross-lingual transfer learning and in-context learning, LLMs are often found to benefit from exposure to non-English languages.
For example,~\citet{turc2021revisiting} find that pretraining one ``pivot'' language could be more effective than English for cross-lingual transfer. 
Similarly, \citet{shi2023language} and \citet{tu2025blessingmultilingualitysystematicanalysis} suggest that greater linguistic diversity in prompting languages improves performance, although~\citet{wang2025lostmultilingualitydissectingcrosslingual} note that multilingual prompting may lead to factual hallucination. 
However, in the context of pruning, it remains unknown whether calibrating on non-English or non-target languages will benefit the performance of pruned mLLMs. 

In short, these studies demonstrate that concepts derived from monolingual settings, particularly those focused on English, often fail to generalize to multilingual contexts involving non-English languages and indicate complex cross-linguistic patterns. Moreover, findings regarding the cross-lingual performance of LLMs are frequently inconsistent across different settings.
Overall, it is difficult to generalize findings from English to non-English scenarios in the context of pruning LLMs. However, most existing work focuses on English-centric settings. Therefore, our work addresses this gap by examining the issue through the lens of language-targeted pruning.

\subsection{Calibration of Post-training Pruning}
Prior research has primarily focused on improving weight importance estimation to optimize pruning performance.
Few studies have examined the role of calibration data, focusing on factors like data quantity~\citep{zhang2024plugandplay} and source~\citep{williams2023doescalibration,bandari-etal-2024-c4}, but these efforts have been confined to English. 

In the multilingual setting, approaches such as multilingual brain surgeon~\citep{zeng-etal-2024-multilingual} aim to retain broad multilingual capabilities by mixing languages in the calibration data according to their pre-training distribution. 
\citet{kim-etal-2024-pruning} leverage the inherent semantic latent space alignment between the pre-training dominant English and non-English languages to improve performance in non-English languages by preserving large-magnitude features that emerge during translation. 
Similarly, \citet{choenni2025mwandaimprovingoneshotpruning} introduce M-Wanda, a multilingual-aware variant of the Wanda algorithm that adjusts pruning based on per-language activation correlations.
However, little is known about how the calibration language influences performance when optimizing for a specific target language rather than preserving general performance across languages. 

Much previous work on multilingual pruning still uses PPL as a first indicator for pruning and calibration performance~\citep{zeng-etal-2024-multilingual,shamrai-2024-language}, despite the limitations of such surface-level metrics in capturing nontrivial reasoning or semantic transfer~\citep{meister-cotterell-2021-language,jaiswal2024compressing}. 
These studies fall short in methodically investigating the extent of these limitations for the language-targeted pruning case we examine.

Our work sheds light on this issue through a multi-view analysis on the calibration language impact on model intrinsics.
Unlike previous studies on multilingual pruning, this work examines language-targeted pruning calibration through a systematic evaluation across multiple LLMs, pruning methods, and downstream tasks.

\section{Methodology}

\begin{table*}[t]
    \centering
    \resizebox{\textwidth}{!}{
\begingroup
\setlength{\tabcolsep}{2pt}
\begin{tabular}{ c c c|ccccccc|ccccccc|ccccccc|}
 &  &  & \multicolumn{7}{c|}{PPL} & \multicolumn{7}{c|}{SNR$_{[dB]}$} & \multicolumn{7}{c|}{Pruning Error $\times 10^3$} \\
 &  &  & AR & DE & EN & ES & RU & SW & ZH & AR & DE & EN & ES & RU & SW & ZH & AR & DE & EN & ES & RU & SW & ZH \\
\hline
\multirow{15}{*}{\rotatebox[origin=c]{90}{Llama-3-8B-Instruct}}
 & - & - & $13.7_{2.1}$ & $10.7_{0.4}$ & $10.2_{0.4}$ & $12.9_{2.2}$ & $9.2_{0.4}$ & $11.6_{0.5}$ & $14.2_{1.2}$ & - & - & - & - & - & - & - & - & - & - & - & - & - & - \\
\cline{2-24}
 & \multirow{7}{*}{\rotatebox[origin=c]{90}{Wanda}}
 & AR & \cellcolor{teal-2!0!white}$30.1_{1.2}$ & \cellcolor{teal-2!71!white}$19.1_{0.8}$ & \cellcolor{teal-2!91!white}$18.6_{0.9}$ & \cellcolor{teal-2!100!white}$23.1_{1.2}$ & \cellcolor{teal-2!50!white}$16.6_{0.6}$ & \cellcolor{teal-2!68!white}$37.9_{2.4}$ & \cellcolor{teal-2!32!white}$33.8_{1.9}$ & \cellcolor{teal-2!0!white}$98.5_{0.3}$ & \cellcolor{teal-2!49!white}$95.3_{0.2}$ & \cellcolor{teal-2!36!white}$95.8_{0.5}$ & \cellcolor{teal-2!43!white}$99.2_{0.3}$ & \cellcolor{teal-2!43!white}$98.2_{0.2}$ & \cellcolor{teal-2!60!white}$84.0_{0.3}$ & \cellcolor{teal-2!34!white}$100.2_{0.7}$ & \cellcolor{teal-2!0!white}$40.1_{0.3}$ & \cellcolor{teal-2!47!white}$43.9_{0.2}$ & \cellcolor{teal-2!34!white}$42.5_{0.5}$ & \cellcolor{teal-2!41!white}$40.3_{0.3}$ & \cellcolor{teal-2!40!white}$40.8_{0.2}$ & \cellcolor{teal-2!56!white}$54.8_{0.4}$ & \cellcolor{teal-2!32!white}$37.6_{0.6}$ \\
\hline
\end{tabular}
\endgroup
    }
    \caption{
        Language-specific perplexity (PPL), Signal-to-noise ratio (SNR) and pruning error at 50\% unstructured sparsity over five pruning runs. 
        The leftmost columns show the model, pruning method, and calibration language (``-'' for unpruned baseline). 
        Columns report performance per evaluation language. 
        The lighter the \textbf{column-wise color-coding}, the better.
        Sub- and superscripts show distances to bootstrapped 95\% confidence intervals. 
        We omit both if negligible or one of them if equal.
    }
    \label{tab:perplexity}
\end{table*}
\begin{table*}[t]
    \centering
    \resizebox{0.98\textwidth}{!}{
\begingroup
\setlength{\tabcolsep}{1.5pt}
\begin{tabular}{ c c c|cccccc|cccccc|cccccc|ccccc|}
 &  &  & \multicolumn{6}{c|}{ARC$_{[acc]}$} & \multicolumn{6}{c|}{Belebele$_{[acc]}$} & \multicolumn{6}{c|}{MMLU$_{[acc]}$} & \multicolumn{5}{c|}{HellaSwag$_{[acc]}$} \\
 &  &  & AR & DE & EN-C & ES & RU & ZH & AR & DE & EN & ES & RU & ZH & AR & DE & EN & ES & RU & ZH & AR & DE & EN & ES & RU \\
\hline
\multirow{15}{*}{\rotatebox[origin=c]{90}{Llama-3-8B-Instruct}}
 & - & - & $29.9_{1.7}$ & $38.1_{1.8}$ & $48.3_{1.9}$ & $39.0_{1.8}$ & $36.4_{1.8}$ & $36.2_{1.8}$ & $45.0_{2.1}$ & $66.1_{2.0}$ & $42.1_{2.1}$ & $72.2_{1.9}$ & $61.0_{2.1}$ & $32.9_{2.1}$ & $25.0_{2.6}$ & $36.1_{2.9}$ & $31.4_{2.8}$ & $43.8_{3.0}$ & $29.1_{2.7}$ & $24.2_{2.6}$ & $36.0_{2.8}$ & $43.6_{2.9}$ & $52.6_{3.0}$ & $45.9_{2.9}$ & $41.6_{2.9}$ \\
\cline{2-26}
 & \multirow{7}{*}{\rotatebox[origin=c]{90}{Wanda}}
 & AR & \cellcolor{teal-2!0!white}$23.0_{2.4}$ & \cellcolor{teal-2!27!white}$29.5_{2.6}$ & \cellcolor{teal-2!27!white}$40.8_{2.7}$ & \cellcolor{teal-2!60!white}$30.6_{2.6}$ & \cellcolor{teal-2!47!white}$28.7_{2.6}$ & \cellcolor{teal-2!0!white}$30.9_{2.6}$ & \cellcolor{teal-2!0!white}$38.9_{2.6}$ & \cellcolor{teal-2!0!white}$47.5_{2.8}$ & \cellcolor{teal-2!60!white}$68.1_{2.5}$ & \cellcolor{teal-2!41!white}$53.4_{2.8}$ & \cellcolor{teal-2!19!white}$49.3_{2.8}$ & \cellcolor{teal-2!11!white}$47.8_{2.8}$ & \cellcolor{teal-2!0!white}$25.9_{1.5}$ & \cellcolor{teal-2!0!white}$29.4_{1.5}$ & \cellcolor{teal-2!24!white}$38.9_{1.6}$ & \cellcolor{teal-2!57!white}$30.5_{1.5}$ & \cellcolor{teal-2!65!white}$25.7_{1.5}$ & \cellcolor{teal-2!17!white}$27.1_{1.5}$ & \cellcolor{teal-2!0!white}$31.2_{1.8}$ & \cellcolor{teal-2!46!white}$36.7_{1.9}$ & \cellcolor{teal-2!59!white}$47.1_{1.9}$ & \cellcolor{teal-2!45!white}$39.7_{1.9}$ & \cellcolor{teal-2!37!white}$35.5_{1.9}$ \\
\hline
\end{tabular}
\endgroup
    }
    \caption{
    Average task accuracy for 50\% unstructured sparsity, computed over five pruning runs. 
    The leftmost columns indicate the model, pruning method, and calibration language (``-'' for unpruned baseline). 
    Columns report accuracy per evaluation language. 
    The lighter the \textbf{column-wise color-coding}, the better.
    Sub- and superscripts show the distance to the bootstrapped 95\% confidence interval (CI). We omit them if negligible or equal.
    For additional results for Llama 3 70B and Aya-23 35B, see Table~\ref{app:table_eval_single_calib_lang_full} in Appendix~\ref{app:supplementary_test_results}.
    }
    \label{tab:eval_single_calib_lang_selection}
\end{table*}

We first compare how calibration languages affect model performance after pruning: using seven distinct calibration languages, we prune a full-sized model to create seven variants. We compare their performance across multiple metrics: pruning error, signal-to-noise ratio, perplexity, and downstream task performance. Implementation details are given in Appendix~\ref{app:implementations}.

\textbf{Models}
Our experiments use two SotA open-source LLM families: Llama-3~\citep{meta2024introducing}, the leading at the time of experiments, and Aya-23~\citep{aryabumi2024aya}, renowned for its multilingual pre-training. 
Given our focus on instructed generation tasks, we use their instruction-tuned versions~\citep{chrysostomou2023lighter}. Our experimental setup includes Llama-3-instruction models in 8B and 70B parameter sizes, alongside Aya-instruction models in 8B and 35B parameter sizes, with the latter moved to the appendix for space constraints.

\textbf{Languages}
We study seven languages: Arabic (AR), German (DE), English (EN), Spanish (ES), Russian (RU), Swahili (SW), and Chinese (ZH). This selection spans six language families and four writing systems and encompasses both high and mid-resource languages. 
To ensure consistent model support, we focus on well-supported languages, while including Swahili as a low-resource outlier, which is not reported to be part of the pre-training data for Llama-3 or Aya-23.
Due to the closed-source nature of pre-training and fine-tuning datasets, language-specific data properties cannot be reliably assessed.
A summary of languages used in our paper is given in Table~\ref{tab:languages} in Appendix \ref{app:language_summary}.

\textbf{Pruning and Calibration} We construct calibration sets for each of the previously listed seven languages from mC4~\citep{2019t5}. Specifically, following~\citet{frantar2023sparsegpt} and~\citet{sun2024a}, we randomly draw 512 sequences of 1,024 tokens for each language from the mC4 test split for that language, ensuring no duplicates by sampling without replacement.
The original Wanda and SparseGPT implementations use 262K calibration tokens. We extend this to 1M tokens for greater sample diversity and potentially more stable pruning, though prior work suggests diminishing returns beyond this size~\citep{sun2024a}.
For mixing different calibration languages, we mix in equal shares and keep the budget of 512 samples.

We focus on two post-training \textit{pruning methods}, Wanda \citep{sun2024a} and SparseGPT \citep{frantar2023sparsegpt}, and prune for 50\% unstructured sparsity, a common setting for LLMs to better observe performance differences while maintaining practical usability~\citep{jaiswal2024compressing}. We keep all other hyperparameters as in the original paper. 

\textbf{Evaluation Downstream Tasks}
We compare the performance of pruned models calibrated on different languages using \textit{perplexity, SNR} and \textit{pruning error} on a subset of the mC4 validation set, and a selection of downstream tasks in different target languages: MKQA~\citep{longpre2020mkqa}, and Belebele~\citep{bandarkar2023belebele}. Additionally, we evaluate on the multilingual mARC, mMMLU and mHellaSwag~\citep{lai2023okapi} datasets as well as their original versions, ARC~\citep{clark2018think}, MMLU~\citep{hendrycks2021measuring} and HellaSwag~\citep{zellers2019hellaswag} in English, due to their lower sensitivity to the choice of calibration samples \citep{williams2023doescalibration}. 
These tasks primarily assess commonsense reasoning, reading comprehension, and question answering using multiple choice questions. We evaluate in a zero-shot setting. Further details of each task are presented in Appendix \ref{app:dataset_summary}. 

\textbf{Uncertainty Estimation}
To estimate uncertainty, we apply non-parametric bootstrapping to compute 95\% confidence intervals~\citep{Efron1993,berg-kirkpatrick-etal-2012-empirical}. For each evaluation configuration, we draw 10,000 bootstrap samples by resampling prediction outcomes with replacement from all test samples across five pruning runs. Each sample includes at least 1,000 instances or 5\% of the total, ensuring sufficient variability. This approach captures empirical uncertainty without assuming any particular distributional form.

\section{Results}\label{sec:results}

\subsection{Pruning Results}\label{sec:subsection_perplexity}

Table \ref{tab:perplexity} evaluates 50\% sparsity pruning performance for calibration on different languages using three common surface-level pruning metrics. 
PPL reflects the general language modeling capability, while SNR and pruning error estimate the deviation of activations from the full-sized model. 
Following \citet{kuzmin2023pruning}, we compute both over all linear sublayers and report size-weighted averages. PPL and SNR, including confidence intervals, are computed globally by aggregating token-level negative log-likelihoods and squared activation differences across five seeds, each using 1,000 samples of 1,024 tokens.

Our findings reveal that no single calibration language consistently outperforms others in perplexity, pruning error, or SNR metrics, despite English-dominated pre-training and unbalanced representation in tokenizer vocabulary.  
The optimal choice depends on the target evaluation language, with \textbf{\textit{calibrating in the target language yielding the lowest magnitude deviation from the full-sized model and lowest perplexity}}.

\subsection{Downstream Task Performance}\label{sec:downstream_performance}

\textit{How to select the calibration language to optimize performance in downstream tasks?}

Table \ref{tab:eval_single_calib_lang_selection} shows the models' performance on downstream tasks. 
Given a downstream task in a specific language, performance is analyzed column-wise among all calibration languages, where lighter colors indicate better results.

First, for both pruning methods, the calibration language affects downstream performance.
For example, when evaluating Llama-3 8B model on Belebele in Spanish, pruning with Wanda results in performance ranging from 47.8 with Swahili calibration to 57.3 with German calibration. In contrast, when pruning Llama-3 8B with SparseGPT the performance on Belebele varies from 35.2 with Chinese calibration to 58.8 with Spanish calibration. 

Unlike the perplexity evaluation in Table \ref{tab:perplexity}, calibration with the target language does not reliably result in good performance, while yielding highly overlapping confidence intervals per pruning configuration and evaluation language. 
For instance, on MMLU, the pruned Llama-3 8B model mainly achieves higher accuracy on evaluation languages other than on its calibration languages. 
For Llama 3 (Aya-23), in 58.3\% (45.8\%) comparison cases, calibration with the target language yields the best performance, 54.2\% (29.2\%) for Wanda and 62.5\% (62.5\%) for SparseGPT (e.g. Russian on ARC and MMLU). 
Comparing calibration with the target language against all other calibrations per pruning configuration and evaluation language yields a global win rate of 66.8\% with 69.9\% (63.8\%) per model, 65.2\% (63\%) for Wanda and 74.6\% (64.5\%) for SparseGPT. 
Therefore, calibrating using the target language often results in acceptable, though not consistently the best performance.
\textbf{\textit{Calibration with the target language benefits but does not guarantee peak performance for reasoning downstream tasks}}. 

Downstream performance for baseline models varies by language due to pertaining distribution and an unbalanced tokenizer. 
English performs best, followed by other Latin-based languages, then Russian. Arabic and Chinese downstream tasks are the most challenging. 
However, despite the dominance of English in pretraining, it is not the optimal calibration language. 
Pruning can alter the original ranking of languages observed in the baseline models. For example, on the Belebele benchmark, the Llama-3 8B model scores $66.1$ for German and $61.0$ for Russian, but the pruned models reverse this trend and achieve a peak accuracy of $47.5$ for German and $51.1$ for Russian. That is, \textbf{\textit{pruning can shift which languages the model performs best or worst on.}}

\textit{Does calibrating on an outlier language or a similar one benefit downstream tasks in non-English?}

Swahili is an out-of-domain language for Llama-3 and Aya-23 as it is not included in the pre-training corpora. In column-wise comparison, the SW cells are among the darkest, indicating that Swahili calibration yields performance among the worst across tasks. 

No clear pattern emerges regarding the similarity of calibration-evaluation language pairs. Latin language pairs such as EN-ES (calibrating in English and evaluating in Spanish) or pairs from the same language family, like EN-DE, do not always yield optimal performance. Conversely, pairs with different writing systems, such as RU-EN or ES-AR, do not consistently perform poorly.
However, calibration with a highly dissimilar language, i.e. Chinese, often results in particularly low accuracy across many tasks and evaluation languages, as demonstrated by the darker row of ``ZH''. In summary, \textbf{\textit{for the limited amount of calibration languages compared, we can not observe an immediate benefit from calibrating with an outlier language or a similar language to the target language.}}

\textit{Does the model or the pruning method matter?}

The results discussed above apply to both Llama-3 and Aya-23 model types as evidenced in Table \ref{app:table_eval_single_calib_lang_full} in Appendix \ref{app:supplementary_test_results} and Table \ref{tab:eval_single_calib_lang_selection}. 

However, between Llama-3 8B and Aya-23 8B, despite their similar decoder-only architecture, there are distinct performance patterns. 
Aya-23 8B generally outperforms the Llama-3 8B model in most evaluation languages and tasks, both before and after pruning. Notably, Aya-23 8B experiences less performance drop after pruning but shows less stable results, often performing better on languages other than the one used for calibration. 

Between Wanda and SparseGPT, Llama-3 8B's and Aya-23 8B's performance degrades less after SparseGPT pruning. 
Our findings highlight the future work required for more robust pruning methods across downstream tasks in different languages.

\begin{table}[t]
    \centering
    \resizebox{\textwidth}{!}{
\begin{tabular}{c c c|c|c|c|c|c|c|}
\hline
 &  &  & \multicolumn{6}{c|}{MKQA$_{[f1]}$}\\
 &  &  & AR & DE & EN & ES & RU & ZH \\
\hline
\multirow{15}{*}{\rotatebox[origin=c]{90}{Llama-3-8B-Instruct}}
 & - & - & 9.1$_{1.6}$ & 27.0$_{2.5}^{2.6}$ & 38.6$_{2.7}$ & 27.4$_{2.5}^{2.6}$ & 16.6$_{2.0}^{2.1}$ & 2.6$_{0.9}^{1.0}$ \\
\cline{2-9}
 & \multirow{7}{*}{\rotatebox[origin=c]{90}{Wanda}}
 & AR & \cellcolor{teal-2!0!white}0.5$_{0.2}$ & \cellcolor{teal-2!74!white}5.7$_{0.6}^{0.7}$ & \cellcolor{teal-2!100!white}18.3$_{1.6}$ & \cellcolor{teal-2!36!white}12.1$_{1.3}^{1.4}$ & \cellcolor{teal-2!39!white}5.3$_{0.9}$ & \cellcolor{teal-2!24!white}2.0$_{0.5}^{0.6}$ \\
 &  & DE & \cellcolor{teal-2!60!white}0.3$_{0.1}^{0.2}$ & \cellcolor{teal-2!54!white}6.0$_{0.7}$ & \cellcolor{teal-2!29!white}19.8$_{1.7}^{1.8}$ & \cellcolor{teal-2!63!white}11.7$_{1.4}$ & \cellcolor{teal-2!0!white}5.9$_{1.0}$ & \cellcolor{teal-2!21!white}2.1$_{0.5}^{0.6}$ \\
 &  & EN & \cellcolor{teal-2!100!white}0.1 & \cellcolor{teal-2!100!white}5.4$_{0.7}$ & \cellcolor{teal-2!0!white}20.4$_{1.7}$ & \cellcolor{teal-2!42!white}12.0$_{1.4}$ & \cellcolor{teal-2!49!white}5.2$_{0.9}$ & \cellcolor{teal-2!62!white}1.5$_{0.4}$ \\
 &  & ES & \cellcolor{teal-2!20!white}0.4$_{0.2}$ & \cellcolor{teal-2!46!white}6.1$_{0.7}^{0.8}$ & \cellcolor{teal-2!10!white}20.2$_{1.7}$ & \cellcolor{teal-2!0!white}12.6$_{1.4}$ & \cellcolor{teal-2!20!white}5.6$_{0.9}^{1.0}$ & \cellcolor{teal-2!100!white}0.9$_{0.3}^{0.4}$ \\
 &  & RU & \cellcolor{teal-2!78!white}0.2 & \cellcolor{teal-2!67!white}5.8$_{0.7}$ & \cellcolor{teal-2!56!white}19.2$_{1.6}^{1.7}$ & \cellcolor{teal-2!15!white}12.4$_{1.3}^{1.4}$ & \cellcolor{teal-2!13!white}5.7$_{1.0}$ & \cellcolor{teal-2!48!white}1.7$_{0.5}$ \\
 &  & SW & \cellcolor{teal-2!3!white}0.5$_{0.2}$ & \cellcolor{teal-2!11!white}6.5$_{0.8}$ & \cellcolor{teal-2!91!white}18.5$_{1.6}^{1.7}$ & \cellcolor{teal-2!100!white}11.2$_{1.3}$ & \cellcolor{teal-2!100!white}4.5$_{0.8}^{0.9}$ & \cellcolor{teal-2!51!white}1.6$_{0.5}$ \\
 &  & ZH & \cellcolor{teal-2!25!white}0.4$_{0.2}$ & \cellcolor{teal-2!0!white}6.6$_{0.8}$ & \cellcolor{teal-2!77!white}18.8$_{1.6}^{1.7}$ & \cellcolor{teal-2!93!white}11.3$_{1.3}^{1.4}$ & \cellcolor{teal-2!96!white}4.6$_{0.8}^{0.9}$ & \cellcolor{teal-2!0!white}2.4$_{0.6}^{0.7}$ \\
\cline{2-9}
 & \multirow{7}{*}{\rotatebox[origin=c]{90}{SparseGPT}}
 & AR & \cellcolor{teal-1!0!white}3.9$_{0.8}$ & \cellcolor{teal-1!98!white}7.6$_{1.0}$ & \cellcolor{teal-1!75!white}20.8$_{1.7}^{1.8}$ & \cellcolor{teal-1!51!white}12.1$_{1.4}^{1.5}$ & \cellcolor{teal-1!46!white}6.8$_{1.1}$ & \cellcolor{teal-1!50!white}1.4$_{0.5}$ \\
 &  & DE & \cellcolor{teal-1!100!white}0.0 & \cellcolor{teal-1!4!white}10.3$_{1.2}$ & \cellcolor{teal-1!76!white}20.8$_{1.7}^{1.8}$ & \cellcolor{teal-1!43!white}12.3$_{1.4}$ & \cellcolor{teal-1!75!white}6.5$_{1.0}$ & \cellcolor{teal-1!26!white}1.6$_{0.5}$ \\
 &  & EN & \cellcolor{teal-1!90!white}0.4$_{0.2}$ & \cellcolor{teal-1!63!white}8.6$_{1.1}$ & \cellcolor{teal-1!0!white}22.4$_{1.8}$ & \cellcolor{teal-1!0!white}13.2$_{1.4}^{1.5}$ & \cellcolor{teal-1!47!white}6.8$_{1.0}^{1.1}$ & \cellcolor{teal-1!0!white}1.8$_{0.5}^{0.6}$ \\
 &  & ES & \cellcolor{teal-1!89!white}0.4$_{0.2}$ & \cellcolor{teal-1!58!white}8.8$_{1.1}$ & \cellcolor{teal-1!52!white}21.3$_{1.7}^{1.8}$ & \cellcolor{teal-1!0!white}13.2$_{1.4}^{1.5}$ & \cellcolor{teal-1!56!white}6.7$_{1.1}$ & \cellcolor{teal-1!40!white}1.5$_{0.5}$ \\
 &  & RU & \cellcolor{teal-1!89!white}0.4$_{0.2}$ & \cellcolor{teal-1!65!white}8.6$_{1.0}^{1.1}$ & \cellcolor{teal-1!95!white}20.4$_{1.7}$ & \cellcolor{teal-1!67!white}11.8$_{1.4}$ & \cellcolor{teal-1!0!white}7.4$_{1.1}$ & \cellcolor{teal-1!43!white}1.5$_{0.5}$ \\
 &  & SW & \cellcolor{teal-1!66!white}1.3$_{0.3}$ & \cellcolor{teal-1!100!white}7.6$_{0.9}^{1.0}$ & \cellcolor{teal-1!78!white}20.7$_{1.7}^{1.8}$ & \cellcolor{teal-1!65!white}11.9$_{1.3}^{1.4}$ & \cellcolor{teal-1!100!white}6.2$_{1.0}$ & \cellcolor{teal-1!100!white}1.1$_{0.4}$ \\
 &  & ZH & \cellcolor{teal-1!77!white}0.9$_{0.2}^{0.3}$ & \cellcolor{teal-1!0!white}10.4$_{1.2}^{1.3}$ & \cellcolor{teal-1!100!white}20.3$_{1.7}$ & \cellcolor{teal-1!100!white}11.2$_{1.3}^{1.4}$ & \cellcolor{teal-1!69!white}6.5$_{1.0}^{1.1}$ & \cellcolor{teal-1!100!white}1.1$_{0.4}^{0.5}$ \\
\hline
\hline
\multirow{15}{*}{\rotatebox[origin=c]{90}{Aya-23-8B}}
 & - & - & 11.2$_{1.5}$ & 20.5$_{1.9}$ & 32.1$_{2.3}^{2.4}$ & 17.9$_{1.7}^{1.8}$ & 13.7$_{1.5}^{1.6}$ & 0.0 \\
\cline{2-9}
 & \multirow{7}{*}{\rotatebox[origin=c]{90}{Wanda}}
 & AR & \cellcolor{navy-2!0!white}6.9$_{0.9}$ & \cellcolor{navy-2!0!white}9.3$_{1.0}$ & \cellcolor{navy-2!75!white}16.4$_{1.3}$ & \cellcolor{navy-2!63!white}10.2$_{1.0}$ & \cellcolor{navy-2!73!white}7.0$_{0.9}$ & \cellcolor{navy-2!61!white}0.1 \\
 &  & DE & \cellcolor{navy-2!44!white}6.2$_{0.8}^{0.9}$ & \cellcolor{navy-2!53!white}8.4$_{0.9}$ & \cellcolor{navy-2!0!white}17.7$_{1.4}$ & \cellcolor{navy-2!39!white}10.6$_{1.0}$ & \cellcolor{navy-2!56!white}7.2$_{0.9}$ & \cellcolor{navy-2!100!white}0.1 \\
 &  & EN & \cellcolor{navy-2!100!white}5.3$_{0.7}^{0.8}$ & \cellcolor{navy-2!53!white}8.4$_{0.9}$ & \cellcolor{navy-2!38!white}17.0$_{1.3}^{1.4}$ & \cellcolor{navy-2!16!white}10.9$_{1.0}^{1.1}$ & \cellcolor{navy-2!27!white}7.6$_{0.9}^{1.0}$ & \cellcolor{navy-2!82!white}0.1 \\
 &  & ES & \cellcolor{navy-2!69!white}5.8$_{0.8}$ & \cellcolor{navy-2!60!white}8.3$_{0.9}$ & \cellcolor{navy-2!29!white}17.2$_{1.3}^{1.4}$ & \cellcolor{navy-2!0!white}11.1$_{1.1}$ & \cellcolor{navy-2!0!white}7.9$_{0.9}^{1.0}$ & \cellcolor{navy-2!100!white}0.1 \\
 &  & RU & \cellcolor{navy-2!67!white}5.9$_{0.8}$ & \cellcolor{navy-2!100!white}7.7$_{0.8}^{0.9}$ & \cellcolor{navy-2!81!white}16.2$_{1.3}^{1.4}$ & \cellcolor{navy-2!40!white}10.6$_{1.0}^{1.1}$ & \cellcolor{navy-2!45!white}7.3$_{0.9}$ & \cellcolor{navy-2!78!white}0.1 \\
 &  & SW & \cellcolor{navy-2!65!white}5.9$_{0.9}$ & \cellcolor{navy-2!55!white}8.4$_{0.9}^{1.0}$ & \cellcolor{navy-2!46!white}16.9$_{1.4}$ & \cellcolor{navy-2!100!white}9.7$_{1.0}$ & \cellcolor{navy-2!100!white}6.7$_{0.9}$ & \cellcolor{navy-2!0!white}0.2$_{0.1}$ \\
 &  & ZH & \cellcolor{navy-2!80!white}5.7$_{0.8}$ & \cellcolor{navy-2!35!white}8.7$_{1.0}$ & \cellcolor{navy-2!100!white}15.9$_{1.3}^{1.4}$ & \cellcolor{navy-2!14!white}10.9$_{1.1}$ & \cellcolor{navy-2!61!white}7.1$_{0.9}$ & \cellcolor{navy-2!65!white}0.1 \\
\cline{2-9}
 & \multirow{7}{*}{\rotatebox[origin=c]{90}{SparseGPT}}
 & AR & \cellcolor{navy-1!0!white}5.8$_{0.8}$ & \cellcolor{navy-1!75!white}8.6$_{0.9}^{1.0}$ & \cellcolor{navy-1!11!white}19.3$_{1.5}^{1.6}$ & \cellcolor{navy-1!40!white}10.3$_{1.0}$ & \cellcolor{navy-1!40!white}6.8$_{0.9}$ & \cellcolor{navy-1!80!white}0.1$_{0.1}^{0.2}$ \\
 &  & DE & \cellcolor{navy-1!63!white}4.9$_{0.7}$ & \cellcolor{navy-1!30!white}8.9$_{0.9}^{1.0}$ & \cellcolor{navy-1!11!white}19.3$_{1.5}$ & \cellcolor{navy-1!40!white}10.3$_{1.0}$ & \cellcolor{navy-1!35!white}6.8$_{0.9}$ & \cellcolor{navy-1!0!white}0.3$_{0.2}^{0.3}$ \\
 &  & EN & \cellcolor{navy-1!77!white}4.7$_{0.7}$ & \cellcolor{navy-1!0!white}9.1$_{0.9}^{1.0}$ & \cellcolor{navy-1!100!white}17.1$_{1.4}$ & \cellcolor{navy-1!33!white}10.4$_{1.0}$ & \cellcolor{navy-1!13!white}7.2$_{0.9}$ & \cellcolor{navy-1!74!white}0.1$_{0.1}^{0.2}$ \\
 &  & ES & \cellcolor{navy-1!13!white}5.6$_{0.8}$ & \cellcolor{navy-1!38!white}8.9$_{1.0}$ & \cellcolor{navy-1!0!white}19.6$_{1.6}$ & \cellcolor{navy-1!91!white}9.8$_{0.9}^{1.0}$ & \cellcolor{navy-1!32!white}6.9$_{0.9}^{1.0}$ & \cellcolor{navy-1!55!white}0.2$_{0.1}^{0.2}$ \\
 &  & RU & \cellcolor{navy-1!44!white}5.2$_{0.7}^{0.8}$ & \cellcolor{navy-1!69!white}8.6$_{1.0}$ & \cellcolor{navy-1!14!white}19.2$_{1.5}^{1.6}$ & \cellcolor{navy-1!0!white}10.8$_{1.0}^{1.1}$ & \cellcolor{navy-1!0!white}7.3$_{0.9}$ & \cellcolor{navy-1!58!white}0.2$_{0.1}^{0.2}$ \\
 &  & SW & \cellcolor{navy-1!78!white}4.7$_{0.7}$ & \cellcolor{navy-1!87!white}8.5$_{1.0}$ & \cellcolor{navy-1!35!white}18.7$_{1.5}$ & \cellcolor{navy-1!73!white}10.0$_{1.0}^{1.1}$ & \cellcolor{navy-1!12!white}7.2$_{1.0}$ & \cellcolor{navy-1!0!white}0.3$_{0.2}^{0.3}$ \\
 &  & ZH & \cellcolor{navy-1!100!white}4.4$_{0.7}$ & \cellcolor{navy-1!100!white}8.4$_{0.9}^{1.0}$ & \cellcolor{navy-1!58!white}18.1$_{1.5}$ & \cellcolor{navy-1!100!white}9.7$_{1.0}$ & \cellcolor{navy-1!100!white}5.9$_{0.8}^{0.9}$ & \cellcolor{navy-1!100!white}0.1 \\
\hline
\end{tabular}
    }
    \caption{Averaged MKQA F1 scores over five pruning runs for the Llama-3 8B and Aya-23 8B models pruned with Wanda and SparseGPT for 50\% unstructured sparsity. Sub- and superscripts show the distance to bootstrapped 95\% confidence intervals. We omit them if negligible or equal.
    }
    \label{tab:eval_mkqa}
\end{table}

\begin{table*}[h]
    \centering
    \resizebox{\textwidth}{!}{
\begingroup
\setlength{\tabcolsep}{2pt}
\begin{tabular}{ c c c|cccccc|cccccc|cccccc|ccccc|}
 &  &  & \multicolumn{6}{c|}{ARC$_{[acc]}$} & \multicolumn{6}{c|}{Belebele$_{[acc]}$} & \multicolumn{6}{c|}{MMLU$_{[acc]}$} & \multicolumn{5}{c|}{HellaSwag$_{[acc]}$} \\
 &  &  & AR & DE & EN-C & ES & RU & ZH & AR & DE & EN & ES & RU & ZH & AR & DE & EN & ES & RU & ZH & AR & DE & EN & ES & RU \\
\hline
\multirow{13}{*}{\rotatebox[origin=c]{90}{Llama-3-8B-Instruct}}
 & - & - & $29.9_{1.7}$ & $38.1_{1.8}$ & $48.3_{1.9}$ & $39.0_{1.8}$ & $36.4_{1.8}$ & $36.2_{1.8}$ & $45.0_{2.1}$ & $66.1_{2.0}$ & $42.1_{2.1}$ & $72.2_{1.9}$ & $61.0_{2.1}$ & $32.9_{2.1}$ & $25.0_{2.6}$ & $36.1_{2.9}$ & $31.4_{2.8}$ & $43.8_{3.0}$ & $29.1_{2.7}$ & $24.2_{2.6}$ & $36.0_{2.8}$ & $43.6_{2.9}$ & $52.6_{3.0}$ & $45.9_{2.9}$ & $41.6_{2.9}$ \\
\cline{2-26}
 & \multirow{12}{*}{\rotatebox[origin=c]{90}{SparseGPT}}
 & ar-de & \cellcolor{teal-1!34!white}$24.8_{2.6}$ & \cellcolor{teal-1!30!white}$31.0_{2.8}$ & \cellcolor{teal-1!74!white}$41.2_{3.0}$ & \cellcolor{teal-1!36!white}$34.2_{2.9}$ & \cellcolor{teal-1!53!white}$30.7_{2.8}$ & \cellcolor{teal-1!61!white}$30.9_{2.9}$ & \cellcolor{teal-1!43!white}$39.0_{2.9}$ & \cellcolor{teal-1!41!white}$50.0_{3.0}$ & \cellcolor{teal-1!37!white}$66.5_{2.9}$ & \cellcolor{teal-1!43!white}$55.3_{2.9}$ & \cellcolor{teal-1!61!white}$45.7_{3.0}$ & \cellcolor{teal-1!60!white}$39.2_{2.9}$ & \cellcolor{teal-1!59!white}$27.0_{0.9}$ & \cellcolor{teal-1!41!white}$34.0_{0.9}$ & \cellcolor{teal-1!52!white}$42.8_{0.9}$ & \cellcolor{teal-1!46!white}$37.0_{0.9}$ & \cellcolor{teal-1!69!white}$30.3_{0.9}$ & \cellcolor{teal-1!79!white}$27.7_{0.9}$ & \cellcolor{teal-1!4!white}$33.1_{1.1}$ & \cellcolor{teal-1!26!white}$39.3_{1.1}$ & \cellcolor{teal-1!48!white}$49.1_{1.1}$ & \cellcolor{teal-1!47!white}$41.8_{1.1}$ & \cellcolor{teal-1!44!white}$37.9_{1.1}$ \\
 &  & ar-en & \cellcolor{teal-1!0!white}$25.5_{2.4}$ & \cellcolor{teal-1!100!white}$29.0_{2.6}$ & \cellcolor{teal-1!54!white}$41.6_{2.8}$ & \cellcolor{teal-1!100!white}$31.7^{2.7}_{2.6}$ & \cellcolor{teal-1!76!white}$30.1_{2.6}$ & \cellcolor{teal-1!51!white}$31.1_{2.6}$ & \cellcolor{teal-1!59!white}$36.8_{2.6}$ & \cellcolor{teal-1!84!white}$44.4_{2.7}$ & \cellcolor{teal-1!95!white}$61.2_{2.7}$ & \cellcolor{teal-1!70!white}$52.4_{2.8}$ & \cellcolor{teal-1!100!white}$41.5_{2.7}$ & \cellcolor{teal-1!79!white}$37.0_{2.6}$ & \cellcolor{teal-1!56!white}$27.2_{1.5}$ & \cellcolor{teal-1!79!white}$31.4_{1.6}$ & \cellcolor{teal-1!77!white}$40.9_{1.6}$ & \cellcolor{teal-1!100!white}$34.3_{1.6}$ & \cellcolor{teal-1!97!white}$28.8_{1.5}$ & \cellcolor{teal-1!94!white}$27.1_{1.5}$ & \cellcolor{teal-1!5!white}$33.1_{1.9}$ & \cellcolor{teal-1!100!white}$38.2_{1.9}$ & \cellcolor{teal-1!12!white}$49.4_{1.9}$ & \cellcolor{teal-1!65!white}$41.6_{1.9}$ & \cellcolor{teal-1!100!white}$37.1_{1.9}$ \\
 &  & ar-es & \cellcolor{teal-1!10!white}$25.3_{2.4}$ & \cellcolor{teal-1!85!white}$29.4_{2.6}$ & \cellcolor{teal-1!85!white}$41.0_{2.8}$ & \cellcolor{teal-1!27!white}$34.5_{2.6}$ & \cellcolor{teal-1!68!white}$30.3_{2.6}$ & \cellcolor{teal-1!74!white}$30.8_{2.6}$ & \cellcolor{teal-1!47!white}$38.4_{2.6}$ & \cellcolor{teal-1!76!white}$45.4_{2.6}$ & \cellcolor{teal-1!100!white}$60.8_{2.7}$ & \cellcolor{teal-1!74!white}$52.1_{2.7}$ & \cellcolor{teal-1!61!white}$45.6_{2.8}$ & \cellcolor{teal-1!72!white}$37.7_{2.7}$ & \cellcolor{teal-1!75!white}$26.2_{1.5}$ & \cellcolor{teal-1!97!white}$30.1_{1.5}$ & \cellcolor{teal-1!100!white}$39.2_{1.6}$ & \cellcolor{teal-1!93!white}$34.7_{1.6}$ & \cellcolor{teal-1!95!white}$28.9_{1.5}$ & \cellcolor{teal-1!100!white}$26.8_{1.5}$ & \cellcolor{teal-1!4!white}$33.2_{1.9}$ & \cellcolor{teal-1!89!white}$38.3_{1.9}$ & \cellcolor{teal-1!82!white}$48.8_{1.9}$ & \cellcolor{teal-1!16!white}$42.2_{1.9}$ & \cellcolor{teal-1!59!white}$37.7_{2.0}$ \\
 &  & ar-ru & \cellcolor{teal-1!16!white}$25.2_{2.4}$ & \cellcolor{teal-1!91!white}$29.3_{2.6}$ & \cellcolor{teal-1!100!white}$40.7_{2.8}$ & \cellcolor{teal-1!81!white}$32.5_{2.7}$ & \cellcolor{teal-1!34!white}$31.3_{2.7}$ & \cellcolor{teal-1!100!white}$30.5_{2.6}$ & \cellcolor{teal-1!0!white}$44.9_{2.8}$ & \cellcolor{teal-1!48!white}$49.2_{2.8}$ & \cellcolor{teal-1!0!white}$69.9_{2.5}$ & \cellcolor{teal-1!60!white}$53.4_{2.7}$ & \cellcolor{teal-1!0!white}$52.4_{2.8}$ & \cellcolor{teal-1!0!white}$46.4_{2.7}$ & \cellcolor{teal-1!0!white}$30.0_{1.5}$ & \cellcolor{teal-1!16!white}$35.8_{1.6}$ & \cellcolor{teal-1!0!white}$46.7_{1.6}$ & \cellcolor{teal-1!43!white}$37.1_{1.6}$ & \cellcolor{teal-1!0!white}$34.1_{1.6}$ & \cellcolor{teal-1!0!white}$31.4_{1.5}$ & \cellcolor{teal-1!0!white}$33.2_{1.9}$ & \cellcolor{teal-1!78!white}$38.5_{1.9}$ & \cellcolor{teal-1!73!white}$48.9_{1.9}$ & \cellcolor{teal-1!100!white}$41.1_{1.9}$ & \cellcolor{teal-1!24!white}$38.2_{1.9}$ \\
 &  & de-en & \cellcolor{teal-1!100!white}$23.4_{2.4}$ & \cellcolor{teal-1!8!white}$31.6_{2.6}$ & \cellcolor{teal-1!0!white}$42.6_{2.9}$ & \cellcolor{teal-1!75!white}$32.7_{2.7}$ & \cellcolor{teal-1!79!white}$30.0_{2.6}$ & \cellcolor{teal-1!9!white}$31.6_{2.7}$ & \cellcolor{teal-1!95!white}$31.9_{2.6}$ & \cellcolor{teal-1!30!white}$51.5_{2.7}$ & \cellcolor{teal-1!0!white}$69.9_{2.5}$ & \cellcolor{teal-1!57!white}$53.8_{2.7}$ & \cellcolor{teal-1!53!white}$46.5_{2.8}$ & \cellcolor{teal-1!19!white}$44.0_{2.7}$ & \cellcolor{teal-1!83!white}$25.8_{1.5}$ & \cellcolor{teal-1!25!white}$35.1_{1.6}$ & \cellcolor{teal-1!20!white}$45.1_{1.7}$ & \cellcolor{teal-1!64!white}$36.1_{1.6}$ & \cellcolor{teal-1!67!white}$30.5_{1.5}$ & \cellcolor{teal-1!41!white}$29.5_{1.5}$ & \cellcolor{teal-1!95!white}$31.4_{1.9}$ & \cellcolor{teal-1!6!white}$39.6_{1.9}$ & \cellcolor{teal-1!0!white}$49.5_{1.9}$ & \cellcolor{teal-1!71!white}$41.5_{1.9}$ & \cellcolor{teal-1!79!white}$37.4_{1.9}$ \\
 &  & de-es & \cellcolor{teal-1!43!white}$24.6_{2.4}$ & \cellcolor{teal-1!0!white}$31.8_{2.6}$ & \cellcolor{teal-1!59!white}$41.5_{2.8}$ & \cellcolor{teal-1!19!white}$34.9_{2.6}$ & \cellcolor{teal-1!56!white}$30.6_{2.6}$ & \cellcolor{teal-1!59!white}$31.0_{2.6}$ & \cellcolor{teal-1!65!white}$36.0_{2.6}$ & \cellcolor{teal-1!36!white}$50.6_{2.7}$ & \cellcolor{teal-1!30!white}$67.1_{2.6}$ & \cellcolor{teal-1!39!white}$55.6_{2.7}$ & \cellcolor{teal-1!50!white}$46.9_{2.7}$ & \cellcolor{teal-1!22!white}$43.7_{2.7}$ & \cellcolor{teal-1!84!white}$25.8_{1.5}$ & \cellcolor{teal-1!35!white}$34.5_{1.6}$ & \cellcolor{teal-1!33!white}$44.2_{1.6}$ & \cellcolor{teal-1!36!white}$37.5_{1.6}$ & \cellcolor{teal-1!71!white}$30.3_{1.5}$ & \cellcolor{teal-1!44!white}$29.4_{1.5}$ & \cellcolor{teal-1!85!white}$31.5_{1.9}$ & \cellcolor{teal-1!9!white}$39.5_{1.9}$ & \cellcolor{teal-1!100!white}$48.6_{1.9}$ & \cellcolor{teal-1!14!white}$42.2_{2.0}$ & \cellcolor{teal-1!57!white}$37.7_{1.9}$ \\
 &  & de-ru & \cellcolor{teal-1!63!white}$24.2_{2.4}$ & \cellcolor{teal-1!12!white}$31.5_{2.6}$ & \cellcolor{teal-1!50!white}$41.7_{2.8}$ & \cellcolor{teal-1!69!white}$32.9_{2.7}$ & \cellcolor{teal-1!0!white}$32.2_{2.7}$ & \cellcolor{teal-1!58!white}$31.0_{2.6}$ & \cellcolor{teal-1!47!white}$38.4_{2.6}$ & \cellcolor{teal-1!40!white}$50.1_{2.7}$ & \cellcolor{teal-1!27!white}$67.4_{2.5}$ & \cellcolor{teal-1!52!white}$54.3_{2.7}$ & \cellcolor{teal-1!42!white}$47.8_{2.8}$ & \cellcolor{teal-1!37!white}$41.9_{2.6}$ & \cellcolor{teal-1!44!white}$27.7_{1.5}$ & \cellcolor{teal-1!0!white}$36.9_{1.6}$ & \cellcolor{teal-1!18!white}$45.3_{1.7}$ & \cellcolor{teal-1!0!white}$39.3_{1.6}$ & \cellcolor{teal-1!24!white}$32.8_{1.6}$ & \cellcolor{teal-1!27!white}$30.1_{1.5}$ & \cellcolor{teal-1!66!white}$31.9_{1.9}$ & \cellcolor{teal-1!0!white}$39.7_{1.9}$ & \cellcolor{teal-1!38!white}$49.1_{1.9}$ & \cellcolor{teal-1!70!white}$41.5_{1.9}$ & \cellcolor{teal-1!0!white}$38.6_{1.9}$ \\
 &  & en-es & \cellcolor{teal-1!95!white}$23.5_{2.4}$ & \cellcolor{teal-1!95!white}$29.2_{2.5}$ & \cellcolor{teal-1!29!white}$42.1_{2.8}$ & \cellcolor{teal-1!9!white}$35.2_{2.6}$ & \cellcolor{teal-1!100!white}$29.4_{2.6}$ & \cellcolor{teal-1!0!white}$31.7_{2.6}$ & \cellcolor{teal-1!100!white}$31.3_{2.6}$ & \cellcolor{teal-1!94!white}$43.1_{2.7}$ & \cellcolor{teal-1!45!white}$65.8_{2.6}$ & \cellcolor{teal-1!40!white}$55.6_{2.8}$ & \cellcolor{teal-1!83!white}$43.3_{2.7}$ & \cellcolor{teal-1!56!white}$39.7_{2.7}$ & \cellcolor{teal-1!100!white}$25.0_{1.5}$ & \cellcolor{teal-1!100!white}$29.9_{1.5}$ & \cellcolor{teal-1!72!white}$41.3_{1.6}$ & \cellcolor{teal-1!64!white}$36.1_{1.6}$ & \cellcolor{teal-1!100!white}$28.7_{1.5}$ & \cellcolor{teal-1!68!white}$28.2_{1.5}$ & \cellcolor{teal-1!100!white}$31.3_{1.9}$ & \cellcolor{teal-1!82!white}$38.4_{1.9}$ & \cellcolor{teal-1!2!white}$49.4_{1.9}$ & \cellcolor{teal-1!5!white}$42.4_{1.9}$ & \cellcolor{teal-1!87!white}$37.3_{1.9}$ \\
 &  & en-ru & \cellcolor{teal-1!76!white}$23.9_{2.4}$ & \cellcolor{teal-1!97!white}$29.1_{2.6}$ & \cellcolor{teal-1!64!white}$41.4_{2.8}$ & \cellcolor{teal-1!80!white}$32.5_{2.7}$ & \cellcolor{teal-1!14!white}$31.8_{2.6}$ & \cellcolor{teal-1!69!white}$30.8_{2.6}$ & \cellcolor{teal-1!96!white}$31.8_{2.5}$ & \cellcolor{teal-1!100!white}$42.4_{2.7}$ & \cellcolor{teal-1!89!white}$61.8_{2.6}$ & \cellcolor{teal-1!100!white}$49.4_{2.7}$ & \cellcolor{teal-1!96!white}$41.9_{2.6}$ & \cellcolor{teal-1!100!white}$34.5^{2.5}_{2.6}$ & \cellcolor{teal-1!70!white}$26.4_{1.5}$ & \cellcolor{teal-1!60!white}$32.7_{1.5}$ & \cellcolor{teal-1!81!white}$40.6_{1.6}$ & \cellcolor{teal-1!55!white}$36.5_{1.6}$ & \cellcolor{teal-1!72!white}$30.2_{1.5}$ & \cellcolor{teal-1!64!white}$28.4_{1.5}$ & \cellcolor{teal-1!71!white}$31.8_{1.9}$ & \cellcolor{teal-1!74!white}$38.5_{1.9}$ & \cellcolor{teal-1!2!white}$49.4_{1.9}$ & \cellcolor{teal-1!86!white}$41.3_{1.9}$ & \cellcolor{teal-1!22!white}$38.2_{1.9}$ \\
 &  & es-ru & \cellcolor{teal-1!83!white}$23.8_{2.4}$ & \cellcolor{teal-1!73!white}$29.8_{2.6}$ & \cellcolor{teal-1!45!white}$41.8_{2.8}$ & \cellcolor{teal-1!0!white}$35.6_{2.7}$ & \cellcolor{teal-1!45!white}$30.9_{2.7}$ & \cellcolor{teal-1!97!white}$30.5_{2.6}$ & \cellcolor{teal-1!46!white}$38.6_{2.7}$ & \cellcolor{teal-1!58!white}$47.9_{2.8}$ & \cellcolor{teal-1!22!white}$67.8_{2.5}$ & \cellcolor{teal-1!39!white}$55.6_{2.8}$ & \cellcolor{teal-1!43!white}$47.7_{2.8}$ & \cellcolor{teal-1!35!white}$42.1_{2.7}$ & \cellcolor{teal-1!60!white}$27.0_{1.5}$ & \cellcolor{teal-1!39!white}$34.2_{1.6}$ & \cellcolor{teal-1!44!white}$43.3_{1.6}$ & \cellcolor{teal-1!19!white}$38.3_{1.6}$ & \cellcolor{teal-1!33!white}$32.3_{1.6}$ & \cellcolor{teal-1!25!white}$30.2_{1.5}$ & \cellcolor{teal-1!66!white}$31.9_{1.9}$ & \cellcolor{teal-1!52!white}$38.9_{1.9}$ & \cellcolor{teal-1!70!white}$48.9_{1.9}$ & \cellcolor{teal-1!0!white}$42.4_{2.0}$ & \cellcolor{teal-1!2!white}$38.5_{1.9}$ \\
 &  & Five langs. & \cellcolor{teal-1!33!white}$24.8_{2.6}$ & \cellcolor{teal-1!40!white}$30.7_{2.8}$ & \cellcolor{teal-1!59!white}$41.5_{3.0}$ & \cellcolor{teal-1!11!white}$35.2_{2.8}$ & \cellcolor{teal-1!47!white}$30.9_{2.8}$ & \cellcolor{teal-1!38!white}$31.2_{2.8}$ & \cellcolor{teal-1!34!white}$40.2_{2.9}$ & \cellcolor{teal-1!31!white}$51.4_{3.0}$ & \cellcolor{teal-1!23!white}$67.8_{2.7}$ & \cellcolor{teal-1!15!white}$58.1_{2.9}$ & \cellcolor{teal-1!43!white}$47.6_{2.9}$ & \cellcolor{teal-1!53!white}$40.1_{2.9}$ & \cellcolor{teal-1!56!white}$27.2_{1.1}$ & \cellcolor{teal-1!36!white}$34.3_{1.1}$ & \cellcolor{teal-1!40!white}$43.6_{1.2}$ & \cellcolor{teal-1!17!white}$38.4_{1.1}$ & \cellcolor{teal-1!58!white}$30.9_{1.1}$ & \cellcolor{teal-1!71!white}$28.1_{1.1}$ & \cellcolor{teal-1!11!white}$33.0_{1.3}$ & \cellcolor{teal-1!31!white}$39.2_{1.4}$ & \cellcolor{teal-1!30!white}$49.2_{1.3}$ & \cellcolor{teal-1!32!white}$42.0_{1.4}$ & \cellcolor{teal-1!28!white}$38.1_{1.4}$ \\
 &  & Seven langs. & \cellcolor{teal-1!25!white}$25.0_{2.4}$ & \cellcolor{teal-1!50!white}$30.4_{2.6}$ & \cellcolor{teal-1!74!white}$41.2_{2.8}$ & \cellcolor{teal-1!24!white}$34.6_{2.7}$ & \cellcolor{teal-1!55!white}$30.7_{2.6}$ & \cellcolor{teal-1!18!white}$31.5_{2.6}$ & \cellcolor{teal-1!20!white}$42.1_{2.6}$ & \cellcolor{teal-1!0!white}$55.5_{2.7}$ & \cellcolor{teal-1!3!white}$69.6_{2.5}$ & \cellcolor{teal-1!0!white}$59.7_{2.7}$ & \cellcolor{teal-1!16!white}$50.6_{2.8}$ & \cellcolor{teal-1!32!white}$42.5_{2.7}$ & \cellcolor{teal-1!40!white}$28.0_{1.5}$ & \cellcolor{teal-1!17!white}$35.7_{1.6}$ & \cellcolor{teal-1!15!white}$45.5_{1.7}$ & \cellcolor{teal-1!0!white}$39.3_{1.7}$ & \cellcolor{teal-1!42!white}$31.8_{1.6}$ & \cellcolor{teal-1!50!white}$29.1_{1.5}$ & \cellcolor{teal-1!15!white}$32.9_{1.9}$ & \cellcolor{teal-1!38!white}$39.1_{1.9}$ & \cellcolor{teal-1!46!white}$49.1_{1.9}$ & \cellcolor{teal-1!38!white}$41.9_{2.0}$ & \cellcolor{teal-1!32!white}$38.1_{1.9}$ \\
\hline
\end{tabular}
\endgroup
    }
    \caption{50\% unstructured sparsity SparseGPT-pruned, mC4-calibrated Llama-3 8B performance averaged over three pruning runs. The lighter, the better. In the row header, the ``Five languages'' refer to  AR, DE, EN, ES, and RU. ``Seven languages'' include AR, DE, EN, ES, RU, SW, ZH. Sub- and superscripts show the distance to bootstrapped 95\% confidence intervals. We omit them if negligible or equal.
    }
    \label{tab:eval_multiligual_calib}
\end{table*}

\subsection{Open Domain Question Answering without Context}\label{sec:open_domain_ansering}

\textit{How does pruning impact the knowledge of LLMs?}

MKQA is a ``closed-book" question-answering task that requires the model to generate answers based solely on its internal knowledge, without external context.\footnote{To ensure fair cross-languages comparisons, the MKQA dataset is fully parallel and primarily consists of entity-based and structured ``atomic'' answer types. See Appendix~\ref{app:dataset_summary} for details.}
As shown in Table \ref{tab:eval_mkqa}, full-size models exhibit significant performance differences across evaluation languages, with Latin languages performing best and Arabic and Chinese performing worst. Pruning leads to a notable accuracy drop across languages, even for English performance. 
In summary, \textbf{\textit{pruning substantially impacts the storage and retrieval of knowledge in a multilingual model across different languages.}}

\subsection{Multiple Calibration Languages}

\textit{Will more calibration languages benefit the downstream?}

We repeat the experiments but include more languages in the calibration set for diversification. 
We experiment with bilingual calibration as well as including all seven languages in the calibration, i.e. multilingual setup. For all setups, the total calibration sample number remains the same, i.e. 512. 

A comparison of Table~\ref{tab:eval_multiligual_calib} with Table~\ref{tab:eval_single_calib_lang_selection} shows that downstream task performance remains similarly unpredictable for bi- or multilingual calibration as it does for monolingual calibration. 
Certain language combinations used for calibration yield good performance among a wider range of target languages, such as AR-DE-EN-ES-RU-SW-ZH on Belebele. % or DE-RU on MMLU. 
On the other hand, calibration sets, such as EN-RU, lead to poor performance across nearly all scenarios. 

In summary, \textbf{\textit{multilingual calibration can help retain performance on a wider range of languages on downstream tasks. However, there is no clear pattern identifying which specific language combinations are most effective for a specific downstream language.}}

\subsection{Impact of Model Sizes}
% by model size ...
To investigate the scaling impact towards pruning behavior, we repeat experiments in Table \ref{app:table_eval_single_calib_lang_full} on the larger Llama-3 70B and Aya-23 35B models, the results of which are reported in Appendix \ref{app:supplementary_test_results}. Overall, pruning performance increases with higher baseline accuracy of the full-sized models. However, we observe that \textbf{\textit{the performance patterns and findings from the smaller models, mentioned above, do not consistently hold true on their bigger counterparts}}. 
For example, while Llama 3 8B exhibits a more predictable diagonal pattern for pruning with SparseGPT, pruning performance for the larger Llama 3 70B depends more on the task than the pruning technique.

\subsection{Quantization}

We further explore the impact of the calibration language in quantization on downstream performance. We use GPTQ~\citep{frantar-gptq} to quantize weights to 4 bits with a group size of 128 and 8 bits with a group size of 128 (equivalent to 50\% sparsity in pruning) on LLaMA-3-8B. We follow our pruning setup for downstream tasks and languages for calibration and evaluation. The results are present in Table~\ref{tab:quantization_w4} and ~\ref{tab:quantization_w8} in Appendix~\ref{app:quantization}, revealing several key findings that are consistent with our findings on pruning: (1) calibrating with the target language yields reasonable although not consistently the best downstream performance; (2) pruning can alter which languages the model performs best or worst on; and (3) calibrating with an outlier or a linguistically similar language does not provide any notable advantage.

\section{Internal Representation Analysis}\label{sec:analysis}

We review Table~\ref{tab:perplexity} and Table~\ref{tab:eval_single_calib_lang_selection} together and find it an interesting pattern that pruning with target-language calibration effectively prunes the less impactful weights, yielding highest SNR and lowest pruning error. 
This benefits simple linguistic tasks like general language modeling captured by PPL, but it is less effective for downstream tasks that require knowledge or reasoning. 
We hypothesize that target-language calibration effectively preserves language-specific features, but not language-agnostic ones, such as knowledge retrieval and reasoning abilities for the pruned multilingual LLMs (see Appendix~\ref{app:section_qualitative_generation_results} for qualitative examples).

To test this, we investigate the internal changes of pruned models at three levels: feature subspace, matrix, and neuron level (columns in a matrix, followed by non-linearity). 
Prior work has separated language-specific features from language-agnostic features at the neuron level~\citep{tang-etal-2024-language,zhao2024large,wang-etal-2024-probing-emergence} or via feature subspace extraction~\citep{xie2022discovering}. We examine how these elements shift post-pruning to explain performance differences.

\subsection{Language-specific Subspace Representations}\label{sec:analysis_subspace}

\begin{figure*}[t]
    \centering
    \begin{subfigure}[b]{\textwidth}
        \centering
        \includegraphics[width=\textwidth]{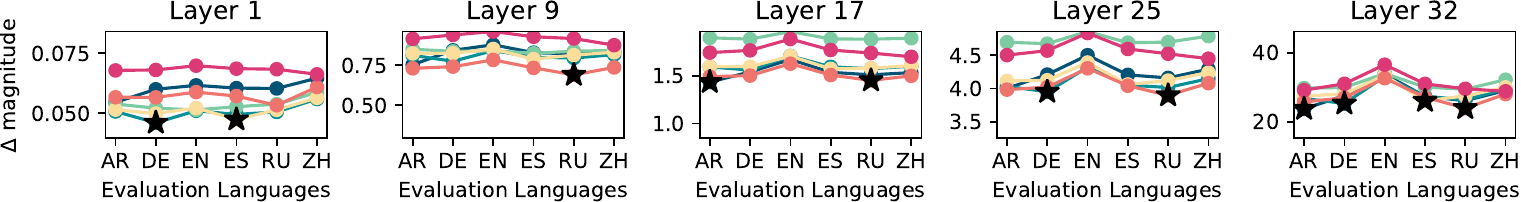}
        \caption{LSAR agnostic features}
        \label{fig:subfig_agnostic_diff}
    \end{subfigure}
    \begin{subfigure}[b]{\textwidth}
        \centering
        \includegraphics[width=\textwidth]{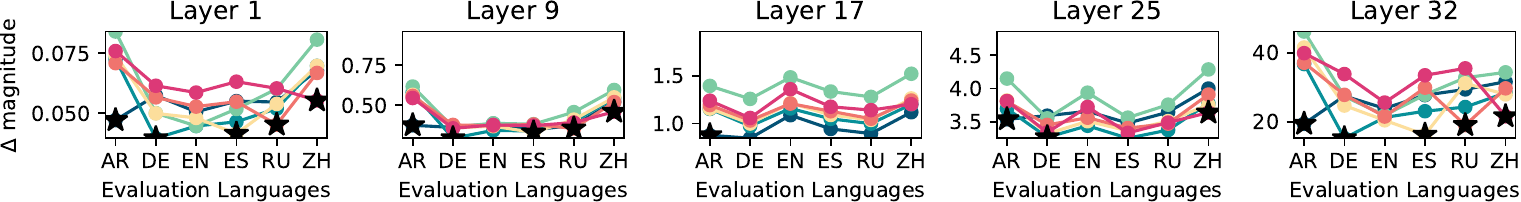}
        \caption{LSAR specific features}
        \label{fig:subfig_specific_diff}
    \end{subfigure}
    \caption{Post-pruning magnitude difference for language-agnostic (Figure~\ref{fig:subfig_agnostic_diff}) and -specific features (Figure~\ref{fig:subfig_specific_diff}), averaged over 900 Belebele samples per language. The X-axis indicates the evaluation language, the calibration language is color-coded: \textcolor{ar}{AR}, \textcolor{de}{DE}, \textcolor{en}{EN}, \textcolor{es}{ES}, \textcolor{ru}{RU}, \textcolor{sw}{ZH}. Larger $\Delta$ means larger pruning error on the respective features. See Figure~\ref{app:figure_lsar_a_mc4_belebele} and~\ref{app:figure_lsar_s_mc4_belebele} in Appendix~\ref{app:further_language_subspace_results} for the full plot over all layers. A star marks matching calibration and evaluation languages with the smallest post-pruning difference.}
    \label{fig:subspace}
\end{figure*}

We use the Low-rank Subspace for language-Agnostic Representations (LSAR) to separate dominant language-specific features from language-agnostic features~\cite{xie2022discovering}. It constructs a mean language embedding matrix $\mathbf{M} \in \mathbb{R}^{d \times L}$ by concatenating averaged language embeddings $\mathbf{e}^{(l)} \in \mathbb{R}^{d}$ for L languages. Subsequently, LSAR decomposes $\mathbf{M}$ into a vector $\boldsymbol\mu$ representing shared signals across languages and a matrix $\mathbf{M}_s$ specifying a low-rank subspace on which different languages express different linguistic signals. This decomposition process is achieved via singular value decomposition on solving: 
\begin{align}
    \min_{\boldsymbol\mu,\mathbf{M}_s,\mathbf{\Gamma}} &\left\| \mathbf{M} - \boldsymbol\mu\mathbb{1}^\top - \mathbf{M}_s\mathbf{\Gamma}^\top \right\|_F^2 \\
    &\text{ s.t. } \boldsymbol\mu \perp \text{Span}(\mathbf{M}_s), \nonumber
\end{align}
with the orthogonality constraint. $\mathbf{\Gamma} \in \mathbb{R}^{L \times r}$ represents the coordinates of language-specific signals along the subspace’s $r$ components, and $\mathbb{1} \in \mathbb{R}^d$ is a vector of all ones. 

Since token-embeddings, through metrics like Pruning Error and SNR, poorly predict downstream task performance, we capture more high-level semantic and syntactic information through prompt-embeddings, layer output embeddings averaged per prompt, excluding special chat-template tokens. 
We extract language-specific features $\mathbf{s}$ from a prompt-embedding $\mathbf{e}$ using $\mathbf{M}_s$ by projecting $\mathbf{e}$ into and back from the low-rank, language signal retaining subspace and obtain the language-agnostic features $\mathbf{a}$ through substraction with $\mathbf{a} = \mathbf{e} - \mathbf{s} = \mathbf{e} - \mathbf{M}_s \mathbf{M}_s^T \mathbf{e}$. 

We use LSAR to decompose the output of each transformer layer of the full-sized and mC4-calibrated, SparseGPT-pruned Llama-3 8B model. 
To evaluate pruning-induced changes, we use all 900 fully-parallel Belebele samples per language, with each sample being semantically identical across languages~\citep{bandarkar2023belebele}. 
For each model $m$, we compute a separate projection matrix $M_s^{(m)}$ with Belebele samples from the six calibration languages, exlcuding SW. 
As we do not require generalization to unseen data as in~\citet{xie2022discovering}, we calculate $M_s^{(m)}$ and evaluate pruning differences using the same 900 Belebele samples per language.

Figure \ref{fig:subspace} shows the layer-wise magnitude of differences ($\Delta$ magnitude) of (a) language-agnostic features and (b) language-specific features after pruning. 
A greater $\Delta$ magnitude suggests greater hidden state changes after pruning, correlating with increased pruning errors. 
Calibrating on the target language reduces pruning errors in language-specific features (Figure~\ref{fig:subfig_specific_diff}), particularly in early, final and selected middle layers as indicated by stars. 
This potentially explains the findings in Section~\ref{sec:subsection_perplexity} that calibration on the target language leads to the lowest perplexity, which is associated with a robust language-specific linguistic modeling capability. On the other hand, as indicated by the relatively flat horizontal lines across languages and layers in Figure~\ref{fig:subfig_agnostic_diff}, the pruning error on the language-agnostic features remains similar regardless of the calibration languages. This pattern helps explain the sub-optimal downstream task performance observed in Table~\ref{app:table_eval_single_calib_lang_full}, where no single calibration language consistently yields optimal performance across downstream tasks, including cases where the calibration is performed on the target language.
That is, \textit{\textbf{the selection of calibration language is unlikely to benefit the language-agnostic features associated with understanding, reasoning, and knowledge retrieval. In contrast, the preservation of language-specific features, such as general language modeling capabilities in early and final layers related to perplexity, depends on the selection of calibration languages.}}

Second, as an overall pruning trend independent of calibration language, the middle layers (as shown in the second to fourth columns in Figure~\ref{fig:subspace}) exhibit greater $\Delta$ magnitude on language-agnostic feature representations and smaller $\Delta$ magnitude on language-specific feature representations. This indicates that pruning errors can be predominantly attributed to the pruning errors on language-agnostic features, with less pruning error arising from language-specific features. Therefore, we conclude \textit{\textbf{pruning affects language-agnostic features, potentially associated with reasoning and knowledge storage, more significantly than it impacts language-specific features.}}

\subsection{Pruning Mask Similarity}\label{sec:analysis_pruning_mask}

To better understand weight pruning decisions based on the calibration set, we conduct a matrix-level analysis by calculating the Intersection over Union (IoU) of pruning masks across different calibration sets, obtaining a measure of pruning mask similarity. 
We use the pruning masks from the Llama-3 8B model pruned with SparseGPT for 50\% unstructured sparsity. To reduce calibration set-dependent noise as prevalent in the downstream tasks, we first compute the intersection $M_I^l$ of pruned neuron indices $M_i^l$ across three pruned models calibrated with different seeds $i$ but in the same language $l$. This intersection represents more stable neuron indices. 

\begin{figure*}[t]
    \begin{subfigure}[b]{\textwidth}
        \centering
        \includegraphics[width=\textwidth]{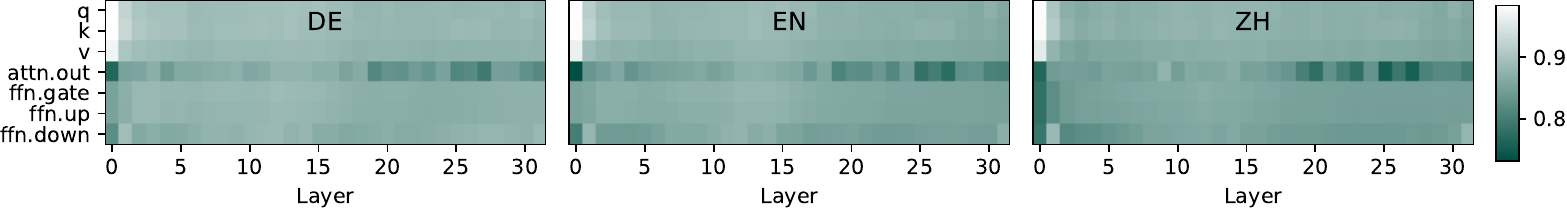}
        \caption{Intra calibration language IOUs}\label{subfig:iou_intra}
    \end{subfigure}
    \vskip 0.1in
    \begin{subfigure}[b]{\textwidth}
        \centering
        \includegraphics[width=\textwidth]{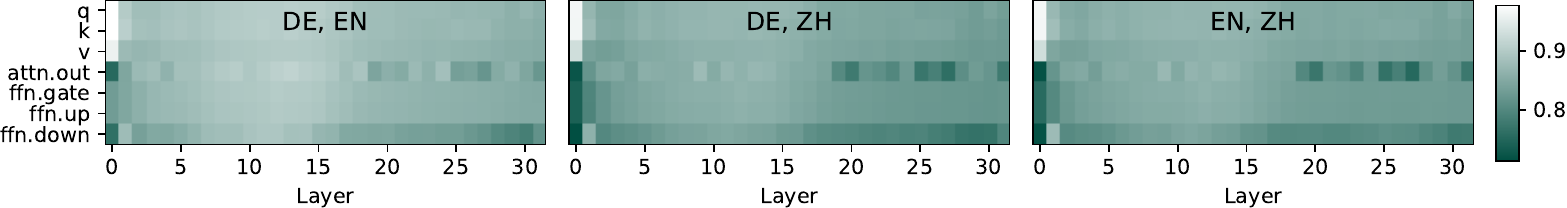}
        \caption{Inter calibration language IOUs}\label{subfig:iou_inter}
    \end{subfigure}
    \caption{Pruning mask similarities (IoUs) between using different calibration languages for 50\% SparseGPT-pruned Llama 3 8B models. \ref{subfig:iou_intra}) IoUs of pruning masks for three calibration sets of the same language. \ref{subfig:iou_inter}) IoUs between pruning masks for different calibration languages. The higher IoU (indicated as a lighter color), the more similar pruning masks between different calibration languages.}
    \label{fig:pruning_mask_similarities}
\end{figure*}

The IoUs in Figure~\ref{subfig:iou_intra} depicts the proportion of $M_I^l$ with respect to all pruned neuron indices. It reveals high pruning mask similarity in the attention query, key and value in the first layer, while the attention output projection varies more significantly, especially after the 20th layer. 
This suggests that \textit{\textbf{pruning struggles to consistently identify essential neurons in the attention output projection, partly responsible for the models reasoning capabilities}}. 

Figure~\ref{subfig:iou_inter} compares the IoU of intersected neuron indices $M_I^l$ from figure~\ref{subfig:iou_intra} across calibration in German, English, and Chinese. Notably, the attention query, key, and value in the first layer consistently achieve high IoUs above 0.95 across all languages, indicating that these components handle inputs similarly, irrespective of language differences.
However, the attention output and FFN down projection show lower IoU, especially in early layers, with similarity peaking in middle layers (3rd to 15th) before decreasing again in later layers. 

This pattern suggests that the attention output and FFN down projection in early and late layers handle language-specific signals, while middle layers process language-agnostic signals, supporting Figure~\ref{fig:subspace}. 
In other words, \textit{\textbf{early layers focus on language comprehension, middle layers on language-independent reasoning, and later layers on generating language-specific predictions. }}
We make similar observations for the Aya-23 8B model as evidenced in Figure~\ref{app:fig_pms_aya-23-8b} of Appendix~\ref{app:pms}. 
This aligns with \citet{zhao2024large}, who propose that LLMs first comprehend queries by converting multilingual inputs into English in the early layers, reason in English in the intermediate layers, and then generate responses aligned with the input language in the final layers. 

\subsection{Language-specific Neurons}\label{sec:analysis_neurons}

This section investigates neuron-level activation frequency changes after pruning using Language Activation Probability Entropy (LAPE) as introduced by \citet{tang-etal-2024-language}. 
We focus on neurons in the up projection of the feed-forward layers (FFNs), followed by the non-linearity. 
A neuron is considered activated when the non-linearity output is greater than zero. LAPE measures the likelihood of individual neurons $i$ of a layer $j$ activating across different language inputs, identifying neurons with high activation probability $p_{i, j}^{(k)}$ for a language $k$ but low probabilities for all others (i.e. low LAPE score) as language-specific. 
Then, neuron-wise LAPE scores with L1 normalized $p_{i, j}^{(k)}$ are computed as: 

\vskip -0.3in
\begin{equation}
    \begin{gathered}
        LAPE_{i, j} = -\sum_{k = 1}^L \widetilde{p}_{i, j}^{(k)} \cdot log( \widetilde{p}_{i, j}^{(k)} ) \\
        \widetilde{\mathbf{p}}_{i, j} = \frac{( p_{i, j}^{(1)}, \dots, p_{i, j}^{(k)}, \dots, p_{i, j}^{(L)} )}{|| ( p_{i, j}^{(1)}, \dots, p_{i, j}^{(k)}, \dots, p_{i, j}^{(L)} ) ||_1}
    \end{gathered}
\end{equation}

Since LAPE scores originally summarize neuron activations across multiple languages, we adapt them into a language-specific metric by correlating them with activation probabilities in a single language. Specifically, we exclude neurons whose activation probability in the target language is lower than the average across all languages, removing those that are language-specific to another language. 
This correlation (below -0.7 Pearson correlation coefficient) not only refines LAPE as a language-specific metric but also enables linking post-pruning changes to language-specific and language-agnostic neuron activation patterns.

\begin{figure*}[t]
    \centering
    \begin{subfigure}[b]{0.48\textwidth}
        \includegraphics[width=2.9in, left]{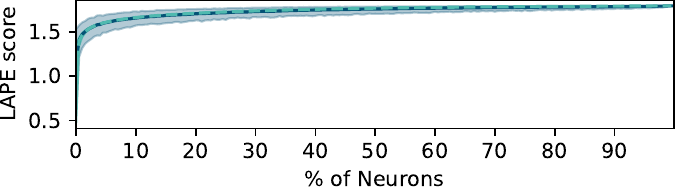}
        \vskip -0.05in
        \caption{LAPE score for full-sized and 50\% pruned model}\label{subfig:lape_score}
        \label{fig:language_entropy_base}
    \end{subfigure}\hfill
    \begin{subfigure}[b]{0.48\textwidth}
        \centering
        \includegraphics[width=2.9in, right]{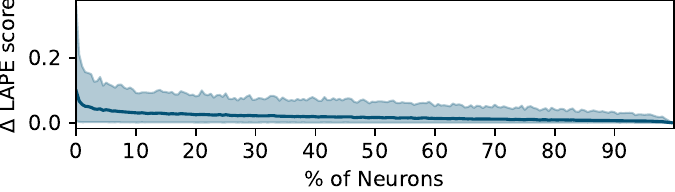}
        \vskip -0.05in
        \caption{Difference of pruned to full-sized model LAPE score}\label{subfig:lape_diff}
    \end{subfigure}
    \vskip 0.1in
    \begin{subfigure}[b]{0.48\textwidth}
        \includegraphics[width=2.9in, left]{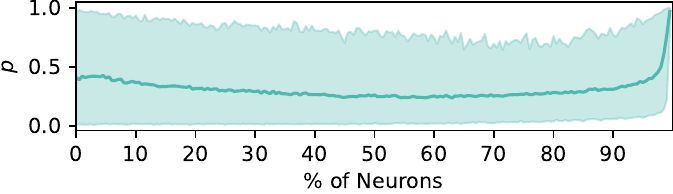}
        \vskip -0.05in
        \caption{Activation probabilities for full-sized model and DE prompts}\label{subfig:lape_act_probs}
    \end{subfigure}\hfill
    \begin{subfigure}[b]{0.48\textwidth}
        \includegraphics[width=2.9in, right]{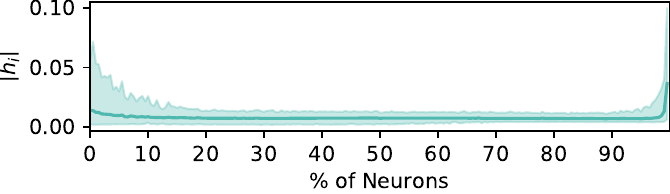}
        \vskip -0.05in
        \caption{Mean activation magnitude for full-sized model and DE prompts}\label{subfig:lape_act_magnitude}
    \end{subfigure}
    \caption{Statistics for FFN neurons of a \textcolor{lape_fs}{full-sized Llama-3 8B} and its \textcolor{lape_pruned}{50\% SparseGPT-pruned} version calibrated for DE. Plots show a 95\% confidence interval and highlighted mean values. 
    All neurons are ordered by ascending LAPE score of the full-sized model as shown by the dashed line in \ref{fig:language_entropy_base}. 
    Additionally, LAPE score and activation probabilities get correlated by removing all neurons with an activation probability in DE that is less then the average activation probability among all languages. 
    \textit{The lower the LAPE score, the more specialized the neuron is for a particular language.} 
    }
    \label{fig:lape_change}
\end{figure*}

Figure~\ref{fig:lape_change} presents LAPE score statistics for the full-sized Llama 3 8B model and its DE-mC4 calibrated, 50\% SparseGPT-pruned version. Activation probabilities were computed from 128 samples per language, each containing 8192 tokens, drawn from the mC4 validation set across the seven previously used calibration languages. 
All neurons are sorted in ascending order based on LAPE scores in the full-sized model correlated with DE, placing language-specific neurons on the left and language-agnostic neurons, i.e. those firing across all languages, on the right. 
Additionally, all metrics are plotted with a 95\% confidence interval and highlighted mean values. 

Figure~\ref{subfig:lape_score} shows a quickly diminishing variance for higher LAPE scores post-pruning. 
This suggests a lower impact of pruning on the activation frequency of high LAPE neurons, i.e. language-agnostic neurons. 
Isolating the LAPE score difference of the pruned to the full-sized model, Figure~\ref{subfig:lape_diff} specifically highlights a \textbf{\textit{strong pruning impact on low LAPE scores, shifting the LAPE score distribution and creating new language-specific (low LAPE) and agnostic (high LAPE) neurons}}. 
In other words, pruning changes activation distributions, potentially causing neurons to activate more or less frequently for a given language.
Such shift in FFNs may contribute to performance degradation in downstream tasks. This hypothesis aligns with previous causal tracing studies, which have identified FFNs are crucial for knowledge retention and retrieval~\citep{meng2022locating, meng2023massediting}.

Moreover, Figure \ref{subfig:lape_act_probs} shows that neurons of lowest LAPE, those most affected by pruning, have low mean activation probabilities compared to neurons with the highest LAPE scores that activate more frequently.
This indicates that \textit{\textbf{pruning struggles to retain the activation frequency of language-specific neurons of low activation probability}}. 
We also examine the average neuron output magnitude for these neglected neurons in Figure~\ref{subfig:lape_act_magnitude}, depicting that despite the low activation probability, low LAPE neurons tend to have high output magnitudes. 
Pruned models show no significant differences in output magnitudes, leading us to conclude that pruning retains average neuron output magnitudes in FFN modules but may fail to preserve activation frequency. 
Figure~\ref{app:fig_lape_aya} and~\ref{app:fig_lape_wanda} in Appendix~\ref{app:lape} confirm the previous findings for the Aya-23 8B model and Wanda-pruning.

\section{Conclusion}

This paper explores how calibration language choice affects pruned multilingual model performance, targeting a specific language for downstream tasks. 
We find that while calibrating on the target language maintains perplexity scores, it does not consistently benefit downstream task performance. 
In fact, calibration in a different language can sometimes yield better results, highlighting limitations in existing pruning strategies. 

Our analysis of internal model representations shows that target language calibration provides limited benefits for features, layers and neurons that encode knowledge and reasoning. 
Current pruning methods prioritize dominant language-specific features, neglecting critical language-agnostic components that are essential for complex tasks. 
This calls for more effective pruning strategies that preserve nuanced language-agnostic information, more reliably identify important neurons across calibration sets with less variance and maintain language-specific activation patterns rather than just focusing on feature magnitudes.

\section*{Acknowledgment}
This research has been funded by the Federal Ministry of Education and research of Germany and the state of North Rhine-Westphalia as part of the Lamarr Institute for Machine Learning and Artificial Intelligence. 

We thank Eryk Majoch\footnote{emajoch1@sheffield.ac.uk}, then an undergraduate in Computer Science at the University of Sheffield, for his valuable contributions to experiments during his internship at the university early in the project.

\bibliography{anthology,tacl2021}
\bibliographystyle{acl_natbib}

\newpage
\onecolumn
\appendix

\section{Implementation Details}\label{app:implementations}
We adopt the code from~\citet{sun2024a} for implementing model pruning. We use EleutherAI Evaluation Harness~\citep{eval-harness} for a robust and reproducible evaluation. We use Huggingface~\citep{wolf-etal-2020-transformers} for loading datasets and models. All experiments are conducted with at most two NVIDIA A100 GPUs.

\section{Limitations} 
\paragraph{Generality of Findings.} Due to resource constraints, we predominantly experimented with the small versions of Llama-3 and Aya-23, and validated our findings with fewer pruning runs on their counterpart large version. Since our results translate between model families, and to bigger model sizes, we assume a certain degree of generalization. Nonetheless, other models trained with different techniques or for other tasks might show different behavior. Given the pace of this research field, it is also unclear whether these results translate to future models. 

\paragraph{Underrepresented Languages.} Our experiments focused on languages with sufficient support for models and downstream tasks. However, this selection does not encompass all languages of interest, particularly mid and low-resource languages that are underrepresented in the pre-training, and challenging to evaluate due to the lack of benchmark support. Future research could benefit from including more languages to explore the interplay between different language families or writing systems and performance after pruning.

\section{Calibration and Test Languages}\label{app:language_summary}

\begin{table}[h]
\resizebox{\textwidth}{!}{%
\begin{tabular}{@{}llllll@{}}
\toprule
\multicolumn{1}{c}{\textbf{Language (Code)}} &
  \multicolumn{1}{c}{\textbf{Language Family}} &
  \multicolumn{1}{c}{\textbf{Writing System}} &
  \multicolumn{1}{c}{\textbf{Script Direction}} &
  \multicolumn{1}{c}{\textbf{Morphological Type}} &
  \multicolumn{1}{c}{\textbf{Geographic Region}} \\ \midrule
Modern Standard Arabic (AR) & Afro-Asiatic   & Arabic         & Right-to-left & Fusional      & Middle East                   \\
German (DE)                 & Germanic       & Latin          & Left-to-right & Fusional      & Western Europe                \\
English (EN)                & Germanic       & Latin          & Left-to-right & Analytic      & Global                        \\
Spanish (ES)                & Romance        & Latin          & Left-to-right & Fusional      & Southern Europe/Latin America \\
Russian (RU)                & Balto-Slavic   & Cyrillic       & Left-to-right & Fusional      & Eastern Europe                \\
Swahili (SW)                & Atlantic-Congo & Latin          & Left-to-right & Agglutinative & East Africa                   \\
Chinese (simplified) (ZH)   & Sino-Tibetan   & Simplified Han & Left-to-right & Isolating     & East Asia                     \\ \bottomrule
\end{tabular}%
}
    \caption{Summary of Languages included in this paper for calibration and evaluation.}
    \label{tab:languages}
\end{table}

\section{Downstream Datasets}\label{app:dataset_summary}

Throughout the paper we used the following widely employed datasets for automated benchmarking. All evaluations were conducted in a zero-shot fashion and employ the chat-template of the respective instruction-tuned model. 

\paragraph{ARC:} The AI2 Reasoning Challenge (ARC) dataset introduced by \citet{clark2018think} tests the reasoning and knowledge capabilities through natural, grad-school multiple choice science questions originally authored for standardized human tests. 
The dataset comprises a total of 7787 questions in english divided into a Challenge set (ARC-C) of hard to answer questions and an Easy set (ARC-E) of questions. 

For evaluation in English, we use the original datasets (e.g. ARC-c \& ARC-e), for all other languages the translated version from ~\citet{lai2023okapi} is utilized.

\paragraph{Belebele:} This carefully curated dataset evaluates 4-way multiple-choice machine reading comprehension among 122 language options, broadly focussing on high-, medium-, and low-resource languages \citep{bandarkar2023belebele}. Each of the 900 samples is based on an English FLORES-200 passage that has been translated into the respective target language by fluent expert speakers. Hence, the dataset is fully parallel, allowing direct performance comparison across all languages. 

\paragraph{HellaSwag:} The HellaSwag dataset by \citet{zellers2019hellaswag} comprises 10042 english samples testing commonsense natural language inference on event descriptions that need to be continued/completed in a multiple-choice fashion. 
Though easily answerable by humans, such paragraph continuation questions still pose a challenge for state-of-the-art LLMs. 

\paragraph{MKQA:} Multilingual Knowledge Questions and Answers (MKQA) \citep{longpre2020mkqa} is an open-domain question-answering evaluation set of 10000 samples aligned across 26 languages by human translators. 
Its question-answer pairs were filtered from the Google Natural Questions dataset \citep{kwiatkowski-etal-2019-natural}, annotating real Google search user questions with answers found on Wikipedia. 
Given a question, the task is to predict the correct answer or give no answer without any additional context provided. Hence, this dataset tests the knowledge retrieval capabilities of models. 
For our evaluation, we remove all unanswerable and questions requiring overly long answers for simplicity, yielding a total 6758 remaining samples. 

There are a total of 6,758 evaluation samples included in our experiment, after eliminating those with unanswerable and long, imprecise answers.

\paragraph{MMLU:} The Massive Multitask Language Understanding (MMLU) dataset \citep{hendrycks2021measuring} is an english benchmark designed to evaluate a model's ability to handle diverse subjects across multiple domains. It contains a total of 14042 question-answer pairs covering 57 task categories, ranging from high school and college-level subjects to professional and specialized knowledge. Each task includes multiple-choice questions, and the dataset measures both the model's factual knowledge and reasoning abilities. 

\paragraph{Translated Datasets from Okapi:} The Okapi framework, introduced by \citet{lai2023okapi}, focuses on instruction tuning LLMs using reinforcement learning from human feedback (RLHF) across multiple languages. As part of its resources, it includes translated versions of the ARC, HellaSwag, and MMLU datasets, generated using ChatGPT. We leverage these translations to complement the evaluation of the original English datasets in multiple languages.

\FloatBarrier
\section{Qualitative Generation Results of Pruned Models}\label{app:section_qualitative_generation_results}

\begin{table}[h!]
    \begin{tabularx}{\textwidth}{X}
        \textbf{Prompt:} Bernice studied some organisms that live together in a field. In which way is a rabbit similar to the plant it eats?
        \\\hline
        \textbf{Answer (EN-calibrated):} A clever question! \textbackslash n\textbackslash nBernice studied organisms that live together in a field, and she found that a rabbit is similar to the plant it eats in the way that both are herbivores. Just like the plant, a rabbit is a herbivore, meaning it eats plants and vegetation.
        \\\hline
        \textbf{Answer (DE-calibrated):} Bernice is studying the organisms that live together in a field! \textbackslash n\textbackslash nIn this case, the rabbit is similar to the plant it eats in the following way:\textbackslash n\textbackslash nBoth the rabbit and the plant are heterotrophs, meaning they are organisms that obtain their nutrients from other organisms. Rabbits are heterotrophic because they eat plants and other animals, whereas plants are autotrophic, meaning they produce their own food from the energy they absorb from the sun. \textbackslash n\textbackslash nIn other words, both the rabbit and the plant are consumers that obtain their energy from sources outside themselves, rather than producing their own food.
    \end{tabularx}
    \caption{Example on the amount of detail retrieved and length of the answer for pruned models.}
\end{table}

\begin{table}[h!]
    \begin{tabularx}{\textwidth}{X}
        \textbf{Prompt:} During which stage of its life cycle does a butterfly spend the most time eating plant \newline leaves?
        \\\hline
        \textbf{Answer (EN-calibrated):} - 
        \\\hline
        \textbf{Answer (DE-calibrated):} Butterflies, also known as Lepidoptera, typically spend the most time eating plant leaves during the larval stage, also known as the caterpillar stage. \textbackslash n\textbackslash nAs a caterpillar, the butterfly spends most of its time feeding on plant leaves, leaves, and other plant material to accumulate energy and nutrients. This feeding period is crucial for the caterpillar's growth and development. \textbackslash n\textbackslash nThe larval stage is typically the longest stage of the butterfly's life cycle, and during this stage, the caterpillar feeds on various plant parts, such as leaves, stems, and flowers, to grow and develop. \textbackslash n\textbackslash nAfter the caterpillar has reached its full size, it forms a pupa or chrysalis, and then emerges as an adult butterfly, ready to start the next generation of its life cycle. \textbackslash n\textbackslash nSo, to answer your question, the stage where a butterfly spends the most time eating plant leaves is the larval stage, which is the caterpillar stage.
    \end{tabularx}
    \caption{Example of one pruned model skipping a prompt answer.}
\end{table}

\begin{table}[h!]
    \begin{tabularx}{\textwidth}{X}
        \textbf{Prompt:} Write me a funny story about a cat and a dog in five sentences.
        \\\hline
        \textbf{Answer (EN-calibrated):} Here's a funny story about a cat and a dog: Whiskers the cat and Rufus the dog were the best of frenemies. They loved to play together, but their playful rivalry was often interrupted by their owners' constant laughter at their silly antics. One day, they decided to team up and steal the family's prized treats from the kitchen. In their excitement, they knocked over a bag of flour, covering themselves in a white cloud. As they ran around the house, they looked like two little snowballs, laughing and chasing each other in circles.
        \\\hline
        \textbf{Answer (DE-calibrated):} One day, a curious cat named Whiskers decided to play a prank on her dog friend, a goofy dog named Rufus. Whiskers snuck into Rufus's doghouse and replaced his dog toys with cat toys, much to Rufus's surprise. Rufus was confused and frustrated when he realized he couldn't find his favorite ball. Whiskers laughed and laughed, thinking it was the best prank she had ever played. Rufus, on the other hand, was left wondering what was going on and why his dog toys were suddenly all cat-related!
    \end{tabularx}
    \vskip 0.2in
    \begin{tabularx}{\textwidth}{X}
        \textbf{Prompt:} Write me a sweet little poem of Alice and her teddy bear in one stanza.
        \\\hline
        \textbf{Answer (EN-calibrated):} Here is a sweet little poem:\textbackslash n Alice holds her teddy bear close,\textbackslash n Her best friend, her heart's gentle rose,\textbackslash n Together they sit, a perfect pair,\textbackslash n A love that's strong, without a care.
        \\\hline
        \textbf{Answer (ZH-calibrated):} Here is a sweet little poem about Alice and her teddy bear:\textbackslash n Alice's teddy bear, her dearest friend,\textbackslash n Lies on her lap, a perfect end.\textbackslash n His button eyes, so bright and wide,\textbackslash n Shining with love, as she snuggles inside.\textbackslash n In his arms, she finds her nest,\textbackslash n A world of dreams, where love is best.\textbackslash n
    \end{tabularx}
    \caption{Examples of language modeling capabilities of 50\% unstructured SparseGPT-pruned models.}
\end{table}

\begin{table}[h!]
    \begin{tabularx}{\textwidth}{X}
        \textbf{Prompt:} Who is the greatest ballerina of all time? Explain why.
        \\\hline
        \textbf{Answer (EN-calibrated):} According to many ballet enthusiasts, the greatest ballerina of all time is considered to be Natalia Makarova.
        \\\hline
        \textbf{Answer (DE-calibrated):} According to many experts and dance historians, the greatest ballerina of all time is Maria Callas (1923-1969). She was a Russian-born American ballet dancer and singer, known for her unique and expressive technique, and her iconic performances in roles like La Scala's 'La Bohème' and 'Eugene Onegin'.
    \end{tabularx}
    \caption{Example of pruned models retrieving different facts (disregarding the actual truth).}
\end{table}

\FloatBarrier
\clearpage
\newpage
\section{Supplementary Test Results}\label{app:supplementary_test_results}

\begin{table*}[h]
    \centering
    \resizebox{\textwidth}{!}{

\begingroup
\begin{tabular}{ c c c|ccccccc|ccccccc|cccccc|ccccc|}
 &  &  & \multicolumn{7}{c|}{ARC$_{[acc]}$} & \multicolumn{7}{c|}{Belebele$_{[acc]}$} & \multicolumn{6}{c|}{MMLU$_{[acc]}$} & \multicolumn{5}{c|}{HellaSwag$_{[acc]}$} \\
 &  &  & AR & DE & EN-E & EN-C & ES & RU & ZH & AR & DE & EN & ES & RU & SW & ZH & AR & DE & EN & ES & RU & ZH & AR & DE & EN & ES & RU \\
\hline
\multirow{15}{*}{\rotatebox[origin=c]{90}{Llama-3-70B-Instruct}}
 & - & - & \cellcolor{teal!0!white}37.0 & \cellcolor{teal!0!white}42.4 & \cellcolor{teal!0!white}75.1 & \cellcolor{teal!0!white}45.1 & \cellcolor{teal!0!white}43.9 & \cellcolor{teal!0!white}38.6 & \cellcolor{teal!0!white}35.9 & \cellcolor{teal!0!white}66.2 & \cellcolor{teal!0!white}75.3 & \cellcolor{teal!0!white}77.8 & \cellcolor{teal!0!white}72.4 & \cellcolor{teal!0!white}71.6 & \cellcolor{teal!0!white}32.3 & \cellcolor{teal!0!white}73.3 & \cellcolor{teal!0!white}44.5 & \cellcolor{teal!0!white}48.7 & \cellcolor{teal!0!white}54.5 & \cellcolor{teal!0!white}50.5 & \cellcolor{teal!0!white}47.5 & \cellcolor{teal!0!white}46.9 & \cellcolor{teal!0!white}41.0 & \cellcolor{teal!0!white}45.7 & \cellcolor{teal!0!white}58.3 & \cellcolor{teal!0!white}49.4 & \cellcolor{teal!0!white}43.5 \\
\cline{2-28}
 & \multirow{7}{*}{\rotatebox[origin=c]{90}{Wanda}}
 & AR & \cellcolor{teal-2!19!white}39.9 & \cellcolor{teal-2!33!white}42.5 & \cellcolor{teal-2!79!white}69.0 & \cellcolor{teal-2!100!white}49.5 & \cellcolor{teal-2!79!white}41.1 & \cellcolor{teal-2!55!white}42.1 & \cellcolor{teal-2!65!white}40.9 & \cellcolor{teal-2!47!white}81.7 & \cellcolor{teal-2!100!white}87.4 & \cellcolor{teal-2!100!white}92.1 & \cellcolor{teal-2!24!white}87.9 & \cellcolor{teal-2!88!white}86.4 & \cellcolor{teal-2!8!white}63.7 & \cellcolor{teal-2!52!white}87.2 & \cellcolor{teal-2!7!white}46.7 & \cellcolor{teal-2!33!white}52.2 & \cellcolor{teal-2!24!white}70.3 & \cellcolor{teal-2!13!white}60.4 & \cellcolor{teal-2!10!white}52.3 & \cellcolor{teal-2!9!white}44.5 & \cellcolor{teal-2!44!white}38.9 & \cellcolor{teal-2!72!white}45.0 & \cellcolor{teal-2!100!white}54.3 & \cellcolor{teal-2!68!white}48.4 & \cellcolor{teal-2!91!white}43.9 \\
 &  & DE & \cellcolor{teal-2!0!white}40.2 & \cellcolor{teal-2!0!white}43.4 & \cellcolor{teal-2!0!white}70.8 & \cellcolor{teal-2!0!white}51.8 & \cellcolor{teal-2!72!white}41.3 & \cellcolor{teal-2!38!white}42.6 & \cellcolor{teal-2!29!white}42.2 & \cellcolor{teal-2!29!white}82.0 & \cellcolor{teal-2!24!white}88.4 & \cellcolor{teal-2!12!white}92.9 & \cellcolor{teal-2!0!white}88.1 & \cellcolor{teal-2!22!white}87.1 & \cellcolor{teal-2!0!white}64.0 & \cellcolor{teal-2!5!white}88.1 & \cellcolor{teal-2!0!white}47.0 & \cellcolor{teal-2!0!white}54.8 & \cellcolor{teal-2!13!white}70.4 & \cellcolor{teal-2!5!white}60.7 & \cellcolor{teal-2!0!white}52.9 & \cellcolor{teal-2!0!white}45.1 & \cellcolor{teal-2!0!white}39.4 & \cellcolor{teal-2!1!white}45.8 & \cellcolor{teal-2!0!white}55.0 & \cellcolor{teal-2!34!white}48.7 & \cellcolor{teal-2!16!white}44.6 \\
 &  & EN & \cellcolor{teal-2!69!white}39.0 & \cellcolor{teal-2!36!white}42.4 & \cellcolor{teal-2!16!white}70.5 & \cellcolor{teal-2!70!white}50.2 & \cellcolor{teal-2!44!white}42.0 & \cellcolor{teal-2!0!white}43.7 & \cellcolor{teal-2!0!white}43.2 & \cellcolor{teal-2!47!white}81.7 & \cellcolor{teal-2!33!white}88.3 & \cellcolor{teal-2!62!white}92.4 & \cellcolor{teal-2!100!white}87.2 & \cellcolor{teal-2!77!white}86.6 & \cellcolor{teal-2!47!white}62.1 & \cellcolor{teal-2!11!white}88.0 & \cellcolor{teal-2!45!white}45.3 & \cellcolor{teal-2!76!white}48.9 & \cellcolor{teal-2!14!white}70.4 & \cellcolor{teal-2!40!white}59.4 & \cellcolor{teal-2!96!white}47.0 & \cellcolor{teal-2!74!white}40.2 & \cellcolor{teal-2!37!white}39.0 & \cellcolor{teal-2!0!white}45.8 & \cellcolor{teal-2!19!white}54.9 & \cellcolor{teal-2!0!white}48.9 & \cellcolor{teal-2!0!white}44.8 \\
 &  & ES & \cellcolor{teal-2!100!white}38.5 & \cellcolor{teal-2!56!white}41.9 & \cellcolor{teal-2!29!white}70.2 & \cellcolor{teal-2!62!white}50.3 & \cellcolor{teal-2!68!white}41.4 & \cellcolor{teal-2!47!white}42.3 & \cellcolor{teal-2!63!white}41.0 & \cellcolor{teal-2!29!white}82.0 & \cellcolor{teal-2!24!white}88.4 & \cellcolor{teal-2!25!white}92.8 & \cellcolor{teal-2!0!white}88.1 & \cellcolor{teal-2!55!white}86.8 & \cellcolor{teal-2!38!white}62.4 & \cellcolor{teal-2!0!white}88.2 & \cellcolor{teal-2!26!white}46.0 & \cellcolor{teal-2!46!white}51.2 & \cellcolor{teal-2!0!white}70.6 & \cellcolor{teal-2!0!white}60.9 & \cellcolor{teal-2!51!white}49.8 & \cellcolor{teal-2!38!white}42.5 & \cellcolor{teal-2!21!white}39.2 & \cellcolor{teal-2!41!white}45.3 & \cellcolor{teal-2!27!white}54.8 & \cellcolor{teal-2!22!white}48.8 & \cellcolor{teal-2!36!white}44.4 \\
 &  & RU & \cellcolor{teal-2!19!white}39.9 & \cellcolor{teal-2!39!white}42.3 & \cellcolor{teal-2!46!white}69.8 & \cellcolor{teal-2!62!white}50.3 & \cellcolor{teal-2!0!white}43.1 & \cellcolor{teal-2!29!white}42.9 & \cellcolor{teal-2!56!white}41.3 & \cellcolor{teal-2!0!white}82.6 & \cellcolor{teal-2!91!white}87.6 & \cellcolor{teal-2!25!white}92.8 & \cellcolor{teal-2!12!white}88.0 & \cellcolor{teal-2!0!white}87.3 & \cellcolor{teal-2!11!white}63.6 & \cellcolor{teal-2!5!white}88.1 & \cellcolor{teal-2!10!white}46.6 & \cellcolor{teal-2!40!white}51.7 & \cellcolor{teal-2!7!white}70.5 & \cellcolor{teal-2!4!white}60.7 & \cellcolor{teal-2!15!white}52.0 & \cellcolor{teal-2!9!white}44.4 & \cellcolor{teal-2!20!white}39.2 & \cellcolor{teal-2!38!white}45.4 & \cellcolor{teal-2!39!white}54.7 & \cellcolor{teal-2!81!white}48.3 & \cellcolor{teal-2!13!white}44.7 \\
 &  & SW & \cellcolor{teal-2!94!white}38.6 & \cellcolor{teal-2!100!white}40.8 & \cellcolor{teal-2!92!white}68.7 & \cellcolor{teal-2!29!white}51.1 & \cellcolor{teal-2!75!white}41.2 & \cellcolor{teal-2!100!white}40.8 & \cellcolor{teal-2!100!white}39.7 & \cellcolor{teal-2!11!white}82.3 & \cellcolor{teal-2!0!white}88.8 & \cellcolor{teal-2!50!white}92.6 & \cellcolor{teal-2!100!white}87.2 & \cellcolor{teal-2!100!white}86.3 & \cellcolor{teal-2!0!white}64.0 & \cellcolor{teal-2!100!white}86.3 & \cellcolor{teal-2!26!white}46.0 & \cellcolor{teal-2!8!white}54.2 & \cellcolor{teal-2!77!white}69.6 & \cellcolor{teal-2!73!white}58.2 & \cellcolor{teal-2!16!white}51.9 & \cellcolor{teal-2!17!white}43.9 & \cellcolor{teal-2!39!white}39.0 & \cellcolor{teal-2!82!white}44.9 & \cellcolor{teal-2!90!white}54.4 & \cellcolor{teal-2!68!white}48.4 & \cellcolor{teal-2!53!white}44.3 \\
 &  & ZH & \cellcolor{teal-2!64!white}39.1 & \cellcolor{teal-2!80!white}41.3 & \cellcolor{teal-2!100!white}68.6 & \cellcolor{teal-2!100!white}49.5 & \cellcolor{teal-2!100!white}40.6 & \cellcolor{teal-2!91!white}41.1 & \cellcolor{teal-2!48!white}41.5 & \cellcolor{teal-2!100!white}80.7 & \cellcolor{teal-2!100!white}87.4 & \cellcolor{teal-2!0!white}93.0 & \cellcolor{teal-2!49!white}87.7 & \cellcolor{teal-2!55!white}86.8 & \cellcolor{teal-2!100!white}60.0 & \cellcolor{teal-2!70!white}86.9 & \cellcolor{teal-2!100!white}43.1 & \cellcolor{teal-2!100!white}47.0 & \cellcolor{teal-2!100!white}69.3 & \cellcolor{teal-2!100!white}57.1 & \cellcolor{teal-2!100!white}46.8 & \cellcolor{teal-2!100!white}38.4 & \cellcolor{teal-2!100!white}38.4 & \cellcolor{teal-2!100!white}44.7 & \cellcolor{teal-2!43!white}54.7 & \cellcolor{teal-2!100!white}48.1 & \cellcolor{teal-2!100!white}43.8 \\
\cline{2-28}
 & \multirow{7}{*}{\rotatebox[origin=c]{90}{SparseGPT}}
 & AR & \cellcolor{teal-1!0!white}40.2 & \cellcolor{teal-1!52!white}42.4 & \cellcolor{teal-1!0!white}72.6 & \cellcolor{teal-1!28!white}51.6 & \cellcolor{teal-1!100!white}41.5 & \cellcolor{teal-1!31!white}43.4 & \cellcolor{teal-1!35!white}41.0 & \cellcolor{teal-1!0!white}81.2 & \cellcolor{teal-1!33!white}88.3 & \cellcolor{teal-1!66!white}92.3 & \cellcolor{teal-1!0!white}88.0 & \cellcolor{teal-1!55!white}84.8 & \cellcolor{teal-1!55!white}63.1 & \cellcolor{teal-1!39!white}86.4 & \cellcolor{teal-1!0!white}46.8 & \cellcolor{teal-1!86!white}43.3 & \cellcolor{teal-1!0!white}71.6 & \cellcolor{teal-1!97!white}57.4 & \cellcolor{teal-1!58!white}46.8 & \cellcolor{teal-1!53!white}38.4 & \cellcolor{teal-1!0!white}39.8 & \cellcolor{teal-1!39!white}46.2 & \cellcolor{teal-1!18!white}56.6 & \cellcolor{teal-1!38!white}49.6 & \cellcolor{teal-1!71!white}44.9 \\
 &  & DE & \cellcolor{teal-1!83!white}38.4 & \cellcolor{teal-1!52!white}42.4 & \cellcolor{teal-1!0!white}72.6 & \cellcolor{teal-1!0!white}52.0 & \cellcolor{teal-1!27!white}42.2 & \cellcolor{teal-1!81!white}42.4 & \cellcolor{teal-1!70!white}40.4 & \cellcolor{teal-1!0!white}81.2 & \cellcolor{teal-1!0!white}89.2 & \cellcolor{teal-1!49!white}92.6 & \cellcolor{teal-1!0!white}88.0 & \cellcolor{teal-1!0!white}86.6 & \cellcolor{teal-1!45!white}63.9 & \cellcolor{teal-1!10!white}87.3 & \cellcolor{teal-1!46!white}45.2 & \cellcolor{teal-1!9!white}50.1 & \cellcolor{teal-1!52!white}70.4 & \cellcolor{teal-1!0!white}60.5 & \cellcolor{teal-1!0!white}50.0 & \cellcolor{teal-1!15!white}42.3 & \cellcolor{teal-1!7!white}39.7 & \cellcolor{teal-1!0!white}47.0 & \cellcolor{teal-1!0!white}56.9 & \cellcolor{teal-1!5!white}50.1 & \cellcolor{teal-1!1!white}45.6 \\
 &  & EN & \cellcolor{teal-1!100!white}38.1 & \cellcolor{teal-1!0!white}43.3 & \cellcolor{teal-1!42!white}71.6 & \cellcolor{teal-1!21!white}51.7 & \cellcolor{teal-1!45!white}42.1 & \cellcolor{teal-1!0!white}44.0 & \cellcolor{teal-1!0!white}41.6 & \cellcolor{teal-1!29!white}80.0 & \cellcolor{teal-1!58!white}87.7 & \cellcolor{teal-1!0!white}93.2 & \cellcolor{teal-1!36!white}87.6 & \cellcolor{teal-1!27!white}85.7 & \cellcolor{teal-1!82!white}61.1 & \cellcolor{teal-1!0!white}87.7 & \cellcolor{teal-1!70!white}44.3 & \cellcolor{teal-1!94!white}42.6 & \cellcolor{teal-1!48!white}70.5 & \cellcolor{teal-1!74!white}58.2 & \cellcolor{teal-1!44!white}47.5 & \cellcolor{teal-1!50!white}38.7 & \cellcolor{teal-1!39!white}39.3 & \cellcolor{teal-1!43!white}46.1 & \cellcolor{teal-1!40!white}56.3 & \cellcolor{teal-1!33!white}49.7 & \cellcolor{teal-1!41!white}45.2 \\
 &  & ES & \cellcolor{teal-1!36!white}39.4 & \cellcolor{teal-1!68!white}42.2 & \cellcolor{teal-1!53!white}71.3 & \cellcolor{teal-1!50!white}51.4 & \cellcolor{teal-1!45!white}42.1 & \cellcolor{teal-1!50!white}43.0 & \cellcolor{teal-1!60!white}40.6 & \cellcolor{teal-1!18!white}80.4 & \cellcolor{teal-1!49!white}87.9 & \cellcolor{teal-1!83!white}92.1 & \cellcolor{teal-1!54!white}87.3 & \cellcolor{teal-1!44!white}85.1 & \cellcolor{teal-1!70!white}62.0 & \cellcolor{teal-1!42!white}86.3 & \cellcolor{teal-1!77!white}44.1 & \cellcolor{teal-1!100!white}42.1 & \cellcolor{teal-1!100!white}69.3 & \cellcolor{teal-1!72!white}58.2 & \cellcolor{teal-1!85!white}45.3 & \cellcolor{teal-1!50!white}38.7 & \cellcolor{teal-1!46!white}39.2 & \cellcolor{teal-1!45!white}46.1 & \cellcolor{teal-1!10!white}56.7 & \cellcolor{teal-1!0!white}50.2 & \cellcolor{teal-1!14!white}45.5 \\
 &  & RU & \cellcolor{teal-1!47!white}39.2 & \cellcolor{teal-1!57!white}42.3 & \cellcolor{teal-1!18!white}72.1 & \cellcolor{teal-1!28!white}51.6 & \cellcolor{teal-1!36!white}42.1 & \cellcolor{teal-1!0!white}44.0 & \cellcolor{teal-1!100!white}39.9 & \cellcolor{teal-1!56!white}78.9 & \cellcolor{teal-1!70!white}87.3 & \cellcolor{teal-1!66!white}92.3 & \cellcolor{teal-1!100!white}86.8 & \cellcolor{teal-1!10!white}86.2 & \cellcolor{teal-1!66!white}62.3 & \cellcolor{teal-1!49!white}86.1 & \cellcolor{teal-1!74!white}44.2 & \cellcolor{teal-1!13!white}49.8 & \cellcolor{teal-1!62!white}70.2 & \cellcolor{teal-1!100!white}57.4 & \cellcolor{teal-1!38!white}47.9 & \cellcolor{teal-1!11!white}42.7 & \cellcolor{teal-1!44!white}39.2 & \cellcolor{teal-1!31!white}46.4 & \cellcolor{teal-1!10!white}56.7 & \cellcolor{teal-1!27!white}49.8 & \cellcolor{teal-1!0!white}45.6 \\
 &  & SW & \cellcolor{teal-1!39!white}39.3 & \cellcolor{teal-1!100!white}41.7 & \cellcolor{teal-1!31!white}71.8 & \cellcolor{teal-1!100!white}50.8 & \cellcolor{teal-1!36!white}42.1 & \cellcolor{teal-1!100!white}42.1 & \cellcolor{teal-1!95!white}40.0 & \cellcolor{teal-1!91!white}77.4 & \cellcolor{teal-1!100!white}86.6 & \cellcolor{teal-1!100!white}91.9 & \cellcolor{teal-1!0!white}88.0 & \cellcolor{teal-1!82!white}83.9 & \cellcolor{teal-1!0!white}67.3 & \cellcolor{teal-1!42!white}86.3 & \cellcolor{teal-1!100!white}43.3 & \cellcolor{teal-1!60!white}45.6 & \cellcolor{teal-1!98!white}69.4 & \cellcolor{teal-1!58!white}58.7 & \cellcolor{teal-1!100!white}44.5 & \cellcolor{teal-1!100!white}33.7 & \cellcolor{teal-1!15!white}39.6 & \cellcolor{teal-1!18!white}46.6 & \cellcolor{teal-1!18!white}56.6 & \cellcolor{teal-1!8!white}50.1 & \cellcolor{teal-1!20!white}45.4 \\
 &  & ZH & \cellcolor{teal-1!87!white}38.3 & \cellcolor{teal-1!52!white}42.4 & \cellcolor{teal-1!100!white}70.3 & \cellcolor{teal-1!100!white}50.8 & \cellcolor{teal-1!0!white}42.5 & \cellcolor{teal-1!90!white}42.3 & \cellcolor{teal-1!4!white}41.5 & \cellcolor{teal-1!100!white}77.1 & \cellcolor{teal-1!74!white}87.2 & \cellcolor{teal-1!91!white}92.0 & \cellcolor{teal-1!45!white}87.4 & \cellcolor{teal-1!100!white}83.3 & \cellcolor{teal-1!100!white}59.8 & \cellcolor{teal-1!100!white}84.6 & \cellcolor{teal-1!68!white}44.4 & \cellcolor{teal-1!0!white}51.0 & \cellcolor{teal-1!58!white}70.3 & \cellcolor{teal-1!56!white}58.7 & \cellcolor{teal-1!27!white}48.5 & \cellcolor{teal-1!0!white}43.9 & \cellcolor{teal-1!100!white}38.5 & \cellcolor{teal-1!100!white}45.0 & \cellcolor{teal-1!100!white}55.4 & \cellcolor{teal-1!100!white}48.7 & \cellcolor{teal-1!100!white}44.6 \\
\hline
\hline
\multirow{15}{*}{\rotatebox[origin=c]{90}{Aya-23-35B}}
 & - & - & \cellcolor{teal!0!white}45.9 & \cellcolor{teal!0!white}46.6 & \cellcolor{teal!0!white}74.7 & \cellcolor{teal!0!white}54.6 & \cellcolor{teal!0!white}47.7 & \cellcolor{teal!0!white}46.3 & \cellcolor{teal!0!white}45.6 & \cellcolor{teal!0!white}86.4 & \cellcolor{teal!0!white}91.4 & \cellcolor{teal!0!white}94.1 & \cellcolor{teal!0!white}90.2 & \cellcolor{teal!0!white}89.2 & \cellcolor{teal!0!white}77.3 & \cellcolor{teal!0!white}89.6 & \cellcolor{teal!0!white}43.3 & \cellcolor{teal!0!white}40.1 & \cellcolor{teal!0!white}76.3 & \cellcolor{teal!0!white}61.6 & \cellcolor{teal!0!white}36.9 & \cellcolor{teal!0!white}28.6 & \cellcolor{teal!0!white}42.6 & \cellcolor{teal!0!white}49.6 & \cellcolor{teal!0!white}59.7 & \cellcolor{teal!0!white}52.6 & \cellcolor{teal!0!white}47.8 \\
\cline{2-28}
 & \multirow{7}{*}{\rotatebox[origin=c]{90}{Wanda}}
 & AR & \cellcolor{navy-2!0!white}41.8 & \cellcolor{navy-2!39!white}46.5 & \cellcolor{navy-2!33!white}80.0 & \cellcolor{navy-2!18!white}54.5 & \cellcolor{navy-2!41!white}48.7 & \cellcolor{navy-2!9!white}44.3 & \cellcolor{navy-2!53!white}40.3 & \cellcolor{navy-2!0!white}80.0 & \cellcolor{navy-2!0!white}81.4 & \cellcolor{navy-2!44!white}87.1 & \cellcolor{navy-2!0!white}82.4 & \cellcolor{navy-2!0!white}78.9 & \cellcolor{navy-2!26!white}39.6 & \cellcolor{navy-2!0!white}81.9 & \cellcolor{navy-2!30!white}48.5 & \cellcolor{navy-2!7!white}53.8 & \cellcolor{navy-2!0!white}60.0 & \cellcolor{navy-2!0!white}55.6 & \cellcolor{navy-2!4!white}51.5 & \cellcolor{navy-2!0!white}52.8 & \cellcolor{navy-2!86!white}42.0 & \cellcolor{navy-2!100!white}47.3 & \cellcolor{navy-2!47!white}61.0 & \cellcolor{navy-2!100!white}52.2 & \cellcolor{navy-2!100!white}45.6 \\
 &  & DE & \cellcolor{navy-2!75!white}39.7 & \cellcolor{navy-2!18!white}47.2 & \cellcolor{navy-2!88!white}79.5 & \cellcolor{navy-2!16!white}54.6 & \cellcolor{navy-2!0!white}49.8 & \cellcolor{navy-2!45!white}43.4 & \cellcolor{navy-2!19!white}41.0 & \cellcolor{navy-2!66!white}77.8 & \cellcolor{navy-2!100!white}78.1 & \cellcolor{navy-2!94!white}86.1 & \cellcolor{navy-2!65!white}80.3 & \cellcolor{navy-2!71!white}76.3 & \cellcolor{navy-2!26!white}39.6 & \cellcolor{navy-2!53!white}81.0 & \cellcolor{navy-2!32!white}48.5 & \cellcolor{navy-2!35!white}53.5 & \cellcolor{navy-2!24!white}59.4 & \cellcolor{navy-2!30!white}55.1 & \cellcolor{navy-2!25!white}51.2 & \cellcolor{navy-2!28!white}52.3 & \cellcolor{navy-2!30!white}42.5 & \cellcolor{navy-2!10!white}48.9 & \cellcolor{navy-2!14!white}61.5 & \cellcolor{navy-2!24!white}53.3 & \cellcolor{navy-2!0!white}46.5 \\
 &  & EN & \cellcolor{navy-2!66!white}39.9 & \cellcolor{navy-2!7!white}47.6 & \cellcolor{navy-2!44!white}79.9 & \cellcolor{navy-2!14!white}54.7 & \cellcolor{navy-2!6!white}49.7 & \cellcolor{navy-2!54!white}43.1 & \cellcolor{navy-2!0!white}41.5 & \cellcolor{navy-2!80!white}77.3 & \cellcolor{navy-2!60!white}79.4 & \cellcolor{navy-2!88!white}86.2 & \cellcolor{navy-2!93!white}79.4 & \cellcolor{navy-2!50!white}77.1 & \cellcolor{navy-2!0!white}40.0 & \cellcolor{navy-2!53!white}81.0 & \cellcolor{navy-2!45!white}48.3 & \cellcolor{navy-2!36!white}53.5 & \cellcolor{navy-2!49!white}58.9 & \cellcolor{navy-2!48!white}54.8 & \cellcolor{navy-2!0!white}51.6 & \cellcolor{navy-2!45!white}51.9 & \cellcolor{navy-2!48!white}42.3 & \cellcolor{navy-2!14!white}48.8 & \cellcolor{navy-2!27!white}61.3 & \cellcolor{navy-2!41!white}53.1 & \cellcolor{navy-2!42!white}46.1 \\
 &  & ES & \cellcolor{navy-2!15!white}41.4 & \cellcolor{navy-2!23!white}47.0 & \cellcolor{navy-2!66!white}79.7 & \cellcolor{navy-2!34!white}53.8 & \cellcolor{navy-2!48!white}48.5 & \cellcolor{navy-2!45!white}43.4 & \cellcolor{navy-2!0!white}41.5 & \cellcolor{navy-2!93!white}76.9 & \cellcolor{navy-2!100!white}78.1 & \cellcolor{navy-2!88!white}86.2 & \cellcolor{navy-2!100!white}79.2 & \cellcolor{navy-2!100!white}75.3 & \cellcolor{navy-2!60!white}39.0 & \cellcolor{navy-2!100!white}80.2 & \cellcolor{navy-2!82!white}47.7 & \cellcolor{navy-2!46!white}53.3 & \cellcolor{navy-2!45!white}59.0 & \cellcolor{navy-2!53!white}54.7 & \cellcolor{navy-2!86!white}50.4 & \cellcolor{navy-2!54!white}51.8 & \cellcolor{navy-2!14!white}42.7 & \cellcolor{navy-2!0!white}49.1 & \cellcolor{navy-2!25!white}61.3 & \cellcolor{navy-2!0!white}53.7 & \cellcolor{navy-2!0!white}46.5 \\
 &  & RU & \cellcolor{navy-2!18!white}41.3 & \cellcolor{navy-2!0!white}47.8 & \cellcolor{navy-2!0!white}80.2 & \cellcolor{navy-2!0!white}55.3 & \cellcolor{navy-2!35!white}48.9 & \cellcolor{navy-2!0!white}44.6 & \cellcolor{navy-2!7!white}41.3 & \cellcolor{navy-2!56!white}78.1 & \cellcolor{navy-2!19!white}80.8 & \cellcolor{navy-2!55!white}86.9 & \cellcolor{navy-2!44!white}81.0 & \cellcolor{navy-2!50!white}77.1 & \cellcolor{navy-2!20!white}39.7 & \cellcolor{navy-2!19!white}81.6 & \cellcolor{navy-2!0!white}49.0 & \cellcolor{navy-2!0!white}53.9 & \cellcolor{navy-2!38!white}59.1 & \cellcolor{navy-2!46!white}54.8 & \cellcolor{navy-2!58!white}50.8 & \cellcolor{navy-2!22!white}52.4 & \cellcolor{navy-2!0!white}42.8 & \cellcolor{navy-2!15!white}48.8 & \cellcolor{navy-2!0!white}61.7 & \cellcolor{navy-2!14!white}53.5 & \cellcolor{navy-2!41!white}46.1 \\
 &  & SW & \cellcolor{navy-2!51!white}40.4 & \cellcolor{navy-2!89!white}44.9 & \cellcolor{navy-2!100!white}79.5 & \cellcolor{navy-2!59!white}52.8 & \cellcolor{navy-2!77!white}47.8 & \cellcolor{navy-2!64!white}42.9 & \cellcolor{navy-2!92!white}39.4 & \cellcolor{navy-2!100!white}76.7 & \cellcolor{navy-2!56!white}79.6 & \cellcolor{navy-2!0!white}88.0 & \cellcolor{navy-2!58!white}80.6 & \cellcolor{navy-2!81!white}76.0 & \cellcolor{navy-2!0!white}40.0 & \cellcolor{navy-2!0!white}81.9 & \cellcolor{navy-2!100!white}47.4 & \cellcolor{navy-2!100!white}52.7 & \cellcolor{navy-2!100!white}57.8 & \cellcolor{navy-2!100!white}54.0 & \cellcolor{navy-2!100!white}50.3 & \cellcolor{navy-2!100!white}50.9 & \cellcolor{navy-2!68!white}42.2 & \cellcolor{navy-2!32!white}48.5 & \cellcolor{navy-2!100!white}60.3 & \cellcolor{navy-2!83!white}52.4 & \cellcolor{navy-2!42!white}46.1 \\
 &  & ZH & \cellcolor{navy-2!100!white}39.0 & \cellcolor{navy-2!100!white}44.6 & \cellcolor{navy-2!94!white}79.5 & \cellcolor{navy-2!100!white}51.1 & \cellcolor{navy-2!100!white}47.2 & \cellcolor{navy-2!100!white}41.9 & \cellcolor{navy-2!100!white}39.2 & \cellcolor{navy-2!76!white}77.4 & \cellcolor{navy-2!93!white}78.3 & \cellcolor{navy-2!100!white}86.0 & \cellcolor{navy-2!48!white}80.9 & \cellcolor{navy-2!34!white}77.7 & \cellcolor{navy-2!100!white}38.3 & \cellcolor{navy-2!73!white}80.7 & \cellcolor{navy-2!75!white}47.8 & \cellcolor{navy-2!90!white}52.8 & \cellcolor{navy-2!38!white}59.1 & \cellcolor{navy-2!87!white}54.2 & \cellcolor{navy-2!41!white}51.0 & \cellcolor{navy-2!41!white}52.0 & \cellcolor{navy-2!100!white}41.8 & \cellcolor{navy-2!53!white}48.1 & \cellcolor{navy-2!100!white}60.3 & \cellcolor{navy-2!87!white}52.4 & \cellcolor{navy-2!92!white}45.6 \\
\cline{2-28}
 & \multirow{7}{*}{\rotatebox[origin=c]{90}{SparseGPT}}
 & AR & \cellcolor{navy-1!31!white}39.9 & \cellcolor{navy-1!41!white}45.2 & \cellcolor{navy-1!55!white}79.0 & \cellcolor{navy-1!25!white}54.0 & \cellcolor{navy-1!0!white}48.8 & \cellcolor{navy-1!82!white}41.7 & \cellcolor{navy-1!40!white}39.6 & \cellcolor{navy-1!22!white}76.8 & \cellcolor{navy-1!25!white}77.4 & \cellcolor{navy-1!11!white}83.1 & \cellcolor{navy-1!4!white}79.4 & \cellcolor{navy-1!9!white}75.9 & \cellcolor{navy-1!0!white}40.3 & \cellcolor{navy-1!17!white}78.9 & \cellcolor{navy-1!16!white}48.4 & \cellcolor{navy-1!23!white}53.6 & \cellcolor{navy-1!0!white}59.0 & \cellcolor{navy-1!14!white}55.0 & \cellcolor{navy-1!0!white}51.8 & \cellcolor{navy-1!28!white}51.1 & \cellcolor{navy-1!65!white}42.4 & \cellcolor{navy-1!100!white}47.2 & \cellcolor{navy-1!55!white}60.5 & \cellcolor{navy-1!81!white}52.0 & \cellcolor{navy-1!72!white}45.7 \\
 &  & DE & \cellcolor{navy-1!68!white}39.3 & \cellcolor{navy-1!31!white}45.4 & \cellcolor{navy-1!58!white}79.0 & \cellcolor{navy-1!85!white}52.6 & \cellcolor{navy-1!30!white}47.8 & \cellcolor{navy-1!82!white}41.7 & \cellcolor{navy-1!40!white}39.6 & \cellcolor{navy-1!43!white}74.4 & \cellcolor{navy-1!49!white}76.1 & \cellcolor{navy-1!15!white}82.9 & \cellcolor{navy-1!16!white}78.3 & \cellcolor{navy-1!17!white}74.9 & \cellcolor{navy-1!96!white}37.4 & \cellcolor{navy-1!48!white}76.3 & \cellcolor{navy-1!35!white}47.9 & \cellcolor{navy-1!1!white}54.0 & \cellcolor{navy-1!38!white}57.6 & \cellcolor{navy-1!13!white}55.1 & \cellcolor{navy-1!23!white}50.8 & \cellcolor{navy-1!27!white}51.1 & \cellcolor{navy-1!66!white}42.4 & \cellcolor{navy-1!26!white}48.5 & \cellcolor{navy-1!39!white}60.7 & \cellcolor{navy-1!43!white}52.7 & \cellcolor{navy-1!44!white}45.9 \\
 &  & EN & \cellcolor{navy-1!59!white}39.4 & \cellcolor{navy-1!41!white}45.2 & \cellcolor{navy-1!20!white}79.5 & \cellcolor{navy-1!81!white}52.7 & \cellcolor{navy-1!10!white}48.5 & \cellcolor{navy-1!28!white}43.0 & \cellcolor{navy-1!50!white}39.4 & \cellcolor{navy-1!29!white}76.0 & \cellcolor{navy-1!15!white}78.0 & \cellcolor{navy-1!0!white}83.9 & \cellcolor{navy-1!21!white}77.9 & \cellcolor{navy-1!8!white}76.0 & \cellcolor{navy-1!3!white}40.2 & \cellcolor{navy-1!22!white}78.4 & \cellcolor{navy-1!46!white}47.5 & \cellcolor{navy-1!25!white}53.5 & \cellcolor{navy-1!51!white}57.1 & \cellcolor{navy-1!5!white}55.3 & \cellcolor{navy-1!23!white}50.8 & \cellcolor{navy-1!29!white}51.1 & \cellcolor{navy-1!84!white}42.2 & \cellcolor{navy-1!0!white}49.0 & \cellcolor{navy-1!24!white}60.8 & \cellcolor{navy-1!23!white}53.0 & \cellcolor{navy-1!1!white}46.3 \\
 &  & ES & \cellcolor{navy-1!63!white}39.3 & \cellcolor{navy-1!3!white}46.1 & \cellcolor{navy-1!17!white}79.6 & \cellcolor{navy-1!37!white}53.8 & \cellcolor{navy-1!41!white}47.4 & \cellcolor{navy-1!7!white}43.5 & \cellcolor{navy-1!30!white}39.7 & \cellcolor{navy-1!66!white}71.9 & \cellcolor{navy-1!64!white}75.2 & \cellcolor{navy-1!38!white}81.3 & \cellcolor{navy-1!12!white}78.7 & \cellcolor{navy-1!46!white}71.3 & \cellcolor{navy-1!92!white}37.6 & \cellcolor{navy-1!37!white}77.2 & \cellcolor{navy-1!48!white}47.5 & \cellcolor{navy-1!12!white}53.8 & \cellcolor{navy-1!0!white}59.1 & \cellcolor{navy-1!0!white}55.5 & \cellcolor{navy-1!38!white}50.2 & \cellcolor{navy-1!43!white}50.6 & \cellcolor{navy-1!7!white}42.9 & \cellcolor{navy-1!33!white}48.4 & \cellcolor{navy-1!7!white}61.0 & \cellcolor{navy-1!0!white}53.5 & \cellcolor{navy-1!0!white}46.3 \\
 &  & RU & \cellcolor{navy-1!0!white}40.5 & \cellcolor{navy-1!0!white}46.2 & \cellcolor{navy-1!0!white}79.8 & \cellcolor{navy-1!0!white}54.6 & \cellcolor{navy-1!30!white}47.8 & \cellcolor{navy-1!0!white}43.7 & \cellcolor{navy-1!0!white}40.3 & \cellcolor{navy-1!0!white}79.2 & \cellcolor{navy-1!0!white}78.9 & \cellcolor{navy-1!13!white}83.0 & \cellcolor{navy-1!0!white}79.9 & \cellcolor{navy-1!12!white}75.4 & \cellcolor{navy-1!3!white}40.2 & \cellcolor{navy-1!26!white}78.1 & \cellcolor{navy-1!0!white}48.9 & \cellcolor{navy-1!0!white}54.0 & \cellcolor{navy-1!23!white}58.2 & \cellcolor{navy-1!13!white}55.1 & \cellcolor{navy-1!20!white}50.9 & \cellcolor{navy-1!0!white}52.0 & \cellcolor{navy-1!0!white}43.0 & \cellcolor{navy-1!9!white}48.8 & \cellcolor{navy-1!0!white}61.1 & \cellcolor{navy-1!29!white}52.9 & \cellcolor{navy-1!0!white}46.3 \\
 &  & SW & \cellcolor{navy-1!31!white}39.9 & \cellcolor{navy-1!96!white}43.8 & \cellcolor{navy-1!100!white}78.4 & \cellcolor{navy-1!100!white}52.3 & \cellcolor{navy-1!53!white}47.0 & \cellcolor{navy-1!96!white}41.4 & \cellcolor{navy-1!100!white}38.5 & \cellcolor{navy-1!100!white}68.2 & \cellcolor{navy-1!100!white}73.2 & \cellcolor{navy-1!100!white}77.2 & \cellcolor{navy-1!100!white}70.4 & \cellcolor{navy-1!100!white}64.8 & \cellcolor{navy-1!51!white}38.8 & \cellcolor{navy-1!100!white}72.0 & \cellcolor{navy-1!100!white}45.9 & \cellcolor{navy-1!100!white}52.1 & \cellcolor{navy-1!100!white}55.4 & \cellcolor{navy-1!100!white}52.4 & \cellcolor{navy-1!100!white}47.7 & \cellcolor{navy-1!100!white}48.9 & \cellcolor{navy-1!83!white}42.2 & \cellcolor{navy-1!67!white}47.8 & \cellcolor{navy-1!100!white}60.0 & \cellcolor{navy-1!91!white}51.8 & \cellcolor{navy-1!94!white}45.6 \\
 &  & ZH & \cellcolor{navy-1!100!white}38.7 & \cellcolor{navy-1!100!white}43.7 & \cellcolor{navy-1!94!white}78.5 & \cellcolor{navy-1!48!white}53.5 & \cellcolor{navy-1!100!white}45.5 & \cellcolor{navy-1!100!white}41.3 & \cellcolor{navy-1!40!white}39.6 & \cellcolor{navy-1!45!white}74.2 & \cellcolor{navy-1!27!white}77.3 & \cellcolor{navy-1!10!white}83.2 & \cellcolor{navy-1!31!white}76.9 & \cellcolor{navy-1!0!white}77.0 & \cellcolor{navy-1!100!white}37.3 & \cellcolor{navy-1!0!white}80.3 & \cellcolor{navy-1!85!white}46.4 & \cellcolor{navy-1!53!white}53.0 & \cellcolor{navy-1!25!white}58.1 & \cellcolor{navy-1!31!white}54.5 & \cellcolor{navy-1!54!white}49.6 & \cellcolor{navy-1!7!white}51.8 & \cellcolor{navy-1!100!white}42.1 & \cellcolor{navy-1!77!white}47.6 & \cellcolor{navy-1!91!white}60.1 & \cellcolor{navy-1!100!white}51.6 & \cellcolor{navy-1!100!white}45.5 \\
\hline
\end{tabular}
\endgroup

    }
    \caption{
    Evaluation of task performance of Llama-3 70B and Aya-23 35B pruned for 50\% unstructured sparsity (128 calibration samples of 8,192 tokens each). 
    The leftmost columns show the model, the pruning technique, and the calibration language used for pruning (one pruning run each). A ``-'' indicates the unpruned reference model. Each column shows the perplexity score of the pruned models on a specific evaluation language. For evaluation in English we use the original datasets (e.g. ARC-c \& ARC-e), for all other languages the translated version from ~\citet{lai2023okapi} is utilized.}
    \label{app:table_eval_single_calib_lang_full}
\end{table*}\label{app:eval_single_calib_lang_full}

\FloatBarrier
\section{Quantization}\label{app:quantization}
\begin{table*}[h]
    \caption{Downstream performance of \textbf{Llama-3 8B} after \textbf{GPTQ} Quantization to 4-bit weights with a group-size of 128  and calibration on \textbf{mC4}.}
    \label{tab:quantization_w4}
    \centering
    \resizebox{\textwidth}{!}{
        \begin{tabular}{ c c c|ccccccc|ccccccc|cccccc|ccccc|}
         &  &  & \multicolumn{7}{c|}{ARC$_{[acc]}$} & \multicolumn{7}{c|}{Belebele$_{[acc]}$} & \multicolumn{6}{c|}{MMLU$_{[acc]}$} & \multicolumn{5}{c|}{HellaSwag$_{[acc]}$} \\
         &  &  & AR & DE & EN-E & EN-C & ES & RU & ZH & AR & DE & EN & ES & RU & SW & ZH & AR & DE & EN & ES & RU & ZH & AR & DE & EN & ES & RU \\
        \hline
        \multirow{7}{*}{\rotatebox[origin=c]{90}{Llama-3-8B-Instruct}}
         & - & - & \cellcolor{teal!0!white}31.7 & \cellcolor{teal!0!white}36.6 & \cellcolor{teal!0!white}75.9 & \cellcolor{teal!0!white}48.5 & \cellcolor{teal!0!white}37.9 & \cellcolor{teal!0!white}36.9 & \cellcolor{teal!0!white}34.8 & \cellcolor{teal!0!white}55.3 & \cellcolor{teal!0!white}69.4 & \cellcolor{teal!0!white}58.2 & \cellcolor{teal!0!white}66.3 & \cellcolor{teal!0!white}62.9 & \cellcolor{teal!0!white}47.3 & \cellcolor{teal!0!white}44.8 & \cellcolor{teal!0!white}27.1 & \cellcolor{teal!0!white}38.2 & \cellcolor{teal!0!white}58.6 & \cellcolor{teal!0!white}43.0 & \cellcolor{teal!0!white}30.3 & \cellcolor{teal!0!white}25.9 & \cellcolor{teal!0!white}36.2 & \cellcolor{teal!0!white}43.2 & \cellcolor{teal!0!white}53.3 & \cellcolor{teal!0!white}46.1 & \cellcolor{teal!0!white}41.7 \\
        \cline{2-28}
         & \multirow{7}{*}{\rotatebox[origin=c]{90}{GPTQ}}
         & AR & \cellcolor{teal!38!white}28.1 & \cellcolor{teal!60!white}32.4 & \cellcolor{teal!56!white}72.2 & \cellcolor{teal!80!white}44.9 & \cellcolor{teal!100!white}33.5 & \cellcolor{teal!39!white}34.0 & \cellcolor{teal!100!white}30.4 & \cellcolor{teal!27!white}47.6 & \cellcolor{teal!9!white}58.4 & \cellcolor{teal!76!white}34.3 & \cellcolor{teal!0!white}65.3 & \cellcolor{teal!50!white}46.7 & \cellcolor{teal!34!white}29.8 & \cellcolor{teal!65!white}28.8 & \cellcolor{teal!12!white}27.2 & \cellcolor{teal!1!white}35.8 & \cellcolor{teal!59!white}55.0 & \cellcolor{teal!0!white}41.2 & \cellcolor{teal!26!white}26.6 & \cellcolor{teal!50!white}23.9 & \cellcolor{teal!0!white}34.6 & \cellcolor{teal!73!white}41.1 & \cellcolor{teal!5!white}52.4 & \cellcolor{teal!2!white}44.5 & \cellcolor{teal!0!white}40.5 \\
         &  & DE & \cellcolor{teal!0!white}29.2 & \cellcolor{teal!0!white}34.8 & \cellcolor{teal!3!white}72.9 & \cellcolor{teal!54!white}45.6 & \cellcolor{teal!9!white}36.0 & \cellcolor{teal!30!white}34.2 & \cellcolor{teal!10!white}32.6 & \cellcolor{teal!72!white}33.6 & \cellcolor{teal!9!white}58.4 & \cellcolor{teal!75!white}34.9 & \cellcolor{teal!30!white}59.2 & \cellcolor{teal!66!white}42.3 & \cellcolor{teal!87!white}24.6 & \cellcolor{teal!78!white}26.9 & \cellcolor{teal!69!white}24.3 & \cellcolor{teal!33!white}32.6 & \cellcolor{teal!93!white}52.8 & \cellcolor{teal!29!white}37.8 & \cellcolor{teal!59!white}25.2 & \cellcolor{teal!61!white}23.7 & \cellcolor{teal!41!white}34.0 & \cellcolor{teal!16!white}41.6 & \cellcolor{teal!17!white}52.3 & \cellcolor{teal!0!white}44.5 & \cellcolor{teal!43!white}40.2 \\
         &  & EN & \cellcolor{teal!76!white}26.9 & \cellcolor{teal!97!white}31.0 & \cellcolor{teal!0!white}72.9 & \cellcolor{teal!41!white}45.9 & \cellcolor{teal!81!white}34.0 & \cellcolor{teal!72!white}33.0 & \cellcolor{teal!21!white}32.3 & \cellcolor{teal!43!white}42.6 & \cellcolor{teal!30!white}52.0 & \cellcolor{teal!41!white}45.7 & \cellcolor{teal!47!white}55.7 & \cellcolor{teal!51!white}46.4 & \cellcolor{teal!84!white}24.9 & \cellcolor{teal!26!white}34.4 & \cellcolor{teal!43!white}25.7 & \cellcolor{teal!38!white}32.2 & \cellcolor{teal!68!white}54.4 & \cellcolor{teal!3!white}40.8 & \cellcolor{teal!32!white}26.4 & \cellcolor{teal!0!white}24.9 & \cellcolor{teal!100!white}33.2 & \cellcolor{teal!100!white}40.9 & \cellcolor{teal!25!white}52.2 & \cellcolor{teal!100!white}44.0 & \cellcolor{teal!61!white}40.1 \\
         &  & ES & \cellcolor{teal!64!white}27.3 & \cellcolor{teal!28!white}33.7 & \cellcolor{teal!43!white}72.3 & \cellcolor{teal!0!white}47.0 & \cellcolor{teal!0!white}36.2 & \cellcolor{teal!0!white}35.1 & \cellcolor{teal!92!white}30.6 & \cellcolor{teal!100!white}24.8 & \cellcolor{teal!100!white}31.3 & \cellcolor{teal!100!white}26.9 & \cellcolor{teal!100!white}45.0 & \cellcolor{teal!100!white}33.0 & \cellcolor{teal!100!white}23.3 & \cellcolor{teal!100!white}23.8 & \cellcolor{teal!100!white}22.8 & \cellcolor{teal!100!white}26.0 & \cellcolor{teal!38!white}56.4 & \cellcolor{teal!59!white}34.3 & \cellcolor{teal!94!white}23.6 & \cellcolor{teal!99!white}22.9 & \cellcolor{teal!39!white}34.0 & \cellcolor{teal!0!white}41.7 & \cellcolor{teal!100!white}51.6 & \cellcolor{teal!65!white}44.2 & \cellcolor{teal!58!white}40.2 \\
         &  & RU & \cellcolor{teal!52!white}27.6 & \cellcolor{teal!73!white}31.9 & \cellcolor{teal!100!white}71.6 & \cellcolor{teal!70!white}45.1 & \cellcolor{teal!68!white}34.4 & \cellcolor{teal!100!white}32.2 & \cellcolor{teal!0!white}32.8 & \cellcolor{teal!81!white}30.7 & \cellcolor{teal!23!white}54.3 & \cellcolor{teal!51!white}42.6 & \cellcolor{teal!51!white}54.8 & \cellcolor{teal!51!white}46.3 & \cellcolor{teal!80!white}25.2 & \cellcolor{teal!54!white}30.3 & \cellcolor{teal!98!white}22.9 & \cellcolor{teal!81!white}27.9 & \cellcolor{teal!100!white}52.4 & \cellcolor{teal!100!white}29.8 & \cellcolor{teal!100!white}23.3 & \cellcolor{teal!100!white}22.9 & \cellcolor{teal!54!white}33.8 & \cellcolor{teal!52!white}41.3 & \cellcolor{teal!22!white}52.3 & \cellcolor{teal!20!white}44.4 & \cellcolor{teal!10!white}40.4 \\
         &  & SW & \cellcolor{teal!100!white}26.3 & \cellcolor{teal!100!white}30.9 & \cellcolor{teal!86!white}71.8 & \cellcolor{teal!100!white}44.4 & \cellcolor{teal!74!white}34.2 & \cellcolor{teal!72!white}33.0 & \cellcolor{teal!14!white}32.5 & \cellcolor{teal!42!white}42.8 & \cellcolor{teal!18!white}55.7 & \cellcolor{teal!27!white}50.3 & \cellcolor{teal!41!white}56.9 & \cellcolor{teal!47!white}47.4 & \cellcolor{teal!0!white}33.1 & \cellcolor{teal!17!white}35.7 & \cellcolor{teal!7!white}27.5 & \cellcolor{teal!0!white}36.0 & \cellcolor{teal!32!white}56.8 & \cellcolor{teal!5!white}40.6 & \cellcolor{teal!17!white}27.1 & \cellcolor{teal!36!white}24.2 & \cellcolor{teal!57!white}33.8 & \cellcolor{teal!55!white}41.2 & \cellcolor{teal!57!white}52.0 & \cellcolor{teal!58!white}44.2 & \cellcolor{teal!100!white}39.9 \\
         &  & ZH & \cellcolor{teal!85!white}26.7 & \cellcolor{teal!15!white}34.2 & \cellcolor{teal!76!white}71.9 & \cellcolor{teal!80!white}44.9 & \cellcolor{teal!12!white}35.9 & \cellcolor{teal!75!white}32.9 & \cellcolor{teal!28!white}32.1 & \cellcolor{teal!0!white}56.3 & \cellcolor{teal!0!white}61.2 & \cellcolor{teal!0!white}59.1 & \cellcolor{teal!43!white}56.4 & \cellcolor{teal!0!white}60.6 & \cellcolor{teal!55!white}27.7 & \cellcolor{teal!0!white}38.2 & \cellcolor{teal!0!white}27.9 & \cellcolor{teal!11!white}34.8 & \cellcolor{teal!0!white}58.8 & \cellcolor{teal!12!white}39.8 & \cellcolor{teal!0!white}27.8 & \cellcolor{teal!0!white}24.9 & \cellcolor{teal!7!white}34.4 & \cellcolor{teal!55!white}41.2 & \cellcolor{teal!0!white}52.5 & \cellcolor{teal!58!white}44.2 & \cellcolor{teal!32!white}40.3 \\
        \cline{2-28}
        \hline
        \end{tabular}
    }
\end{table*}

\begin{table*}[!htbp]
    \caption{Downstream performance of \textbf{Llama-3 8B} after \textbf{GPTQ} Quantization to 8-bit weights with a group-size of 128 and calibration on \textbf{mC4}.}
    \label{tab:quantization_w8}
    \centering
    \resizebox{\textwidth}{!}{
        \begin{tabular}{ c c c|ccccccc|ccccccc|cccccc|ccccc|}
         &  &  & \multicolumn{7}{c|}{ARC$_{[acc]}$} & \multicolumn{7}{c|}{Belebele$_{[acc]}$} & \multicolumn{6}{c|}{MMLU$_{[acc]}$} & \multicolumn{5}{c|}{HellaSwag$_{[acc]}$} \\
         &  &  & AR & DE & EN-E & EN-C & ES & RU & ZH & AR & DE & EN & ES & RU & SW & ZH & AR & DE & EN & ES & RU & ZH & AR & DE & EN & ES & RU \\
        \hline
        \multirow{7}{*}{\rotatebox[origin=c]{90}{Llama-3-8B-Instruct}}
         & - & - & \cellcolor{teal!0!white}31.7 & \cellcolor{teal!0!white}36.6 & \cellcolor{teal!0!white}75.9 & \cellcolor{teal!0!white}48.5 & \cellcolor{teal!0!white}37.9 & \cellcolor{teal!0!white}36.9 & \cellcolor{teal!0!white}34.8 & \cellcolor{teal!0!white}55.3 & \cellcolor{teal!0!white}69.4 & \cellcolor{teal!0!white}58.2 & \cellcolor{teal!0!white}66.3 & \cellcolor{teal!0!white}62.9 & \cellcolor{teal!0!white}47.3 & \cellcolor{teal!0!white}44.8 & \cellcolor{teal!0!white}27.1 & \cellcolor{teal!0!white}38.2 & \cellcolor{teal!0!white}58.6 & \cellcolor{teal!0!white}43.0 & \cellcolor{teal!0!white}30.3 & \cellcolor{teal!0!white}25.9 & \cellcolor{teal!0!white}36.2 & \cellcolor{teal!0!white}43.2 & \cellcolor{teal!0!white}53.3 & \cellcolor{teal!0!white}46.1 & \cellcolor{teal!0!white}41.7 \\
        \cline{2-28}
         & \multirow{7}{*}{\rotatebox[origin=c]{90}{GPTQ}}
         & AR & \cellcolor{teal!50!white}31.3 & \cellcolor{teal!14!white}36.9 & \cellcolor{teal!40!white}75.8 & \cellcolor{teal!75!white}48.4 & \cellcolor{teal!83!white}38.0 & \cellcolor{teal!33!white}37.3 & \cellcolor{teal!100!white}35.1 & \cellcolor{teal!79!white}56.1 & \cellcolor{teal!60!white}70.2 & \cellcolor{teal!57!white}60.1 & \cellcolor{teal!6!white}68.3 & \cellcolor{teal!49!white}62.9 & \cellcolor{teal!49!white}47.7 & \cellcolor{teal!14!white}48.2 & \cellcolor{teal!0!white}27.6 & \cellcolor{teal!17!white}38.6 & \cellcolor{teal!47!white}59.0 & \cellcolor{teal!44!white}43.6 & \cellcolor{teal!12!white}30.9 & \cellcolor{teal!17!white}26.3 & \cellcolor{teal!28!white}36.2 & \cellcolor{teal!100!white}43.2 & \cellcolor{teal!24!white}53.4 & \cellcolor{teal!77!white}46.1 & \cellcolor{teal!43!white}41.7 \\
         &  & DE & \cellcolor{teal!0!white}31.5 & \cellcolor{teal!28!white}36.8 & \cellcolor{teal!0!white}75.9 & \cellcolor{teal!50!white}48.5 & \cellcolor{teal!100!white}37.9 & \cellcolor{teal!100!white}36.8 & \cellcolor{teal!0!white}35.6 & \cellcolor{teal!0!white}57.0 & \cellcolor{teal!0!white}70.6 & \cellcolor{teal!50!white}60.2 & \cellcolor{teal!13!white}68.2 & \cellcolor{teal!60!white}62.8 & \cellcolor{teal!0!white}48.3 & \cellcolor{teal!28!white}47.9 & \cellcolor{teal!0!white}27.6 & \cellcolor{teal!0!white}38.7 & \cellcolor{teal!35!white}59.1 & \cellcolor{teal!18!white}43.9 & \cellcolor{teal!3!white}31.0 & \cellcolor{teal!9!white}26.3 & \cellcolor{teal!100!white}36.2 & \cellcolor{teal!24!white}43.3 & \cellcolor{teal!100!white}53.3 & \cellcolor{teal!100!white}46.0 & \cellcolor{teal!18!white}41.7 \\
         &  & EN & \cellcolor{teal!25!white}31.4 & \cellcolor{teal!100!white}36.4 & \cellcolor{teal!19!white}75.9 & \cellcolor{teal!0!white}48.6 & \cellcolor{teal!33!white}38.3 & \cellcolor{teal!44!white}37.2 & \cellcolor{teal!66!white}35.3 & \cellcolor{teal!100!white}55.9 & \cellcolor{teal!100!white}70.0 & \cellcolor{teal!100!white}59.4 & \cellcolor{teal!100!white}66.8 & \cellcolor{teal!100!white}62.3 & \cellcolor{teal!100!white}47.0 & \cellcolor{teal!100!white}46.2 & \cellcolor{teal!100!white}27.4 & \cellcolor{teal!100!white}38.1 & \cellcolor{teal!32!white}59.1 & \cellcolor{teal!100!white}43.0 & \cellcolor{teal!100!white}30.5 & \cellcolor{teal!100!white}26.0 & \cellcolor{teal!42!white}36.2 & \cellcolor{teal!66!white}43.2 & \cellcolor{teal!49!white}53.3 & \cellcolor{teal!77!white}46.1 & \cellcolor{teal!100!white}41.6 \\
         &  & ES & \cellcolor{teal!100!white}31.1 & \cellcolor{teal!0!white}37.0 & \cellcolor{teal!100!white}75.7 & \cellcolor{teal!50!white}48.5 & \cellcolor{teal!83!white}38.0 & \cellcolor{teal!55!white}37.1 & \cellcolor{teal!83!white}35.2 & \cellcolor{teal!9!white}56.9 & \cellcolor{teal!20!white}70.4 & \cellcolor{teal!28!white}60.6 & \cellcolor{teal!26!white}68.0 & \cellcolor{teal!40!white}63.0 & \cellcolor{teal!33!white}47.9 & \cellcolor{teal!0!white}48.6 & \cellcolor{teal!24!white}27.6 & \cellcolor{teal!78!white}38.2 & \cellcolor{teal!100!white}58.7 & \cellcolor{teal!42!white}43.6 & \cellcolor{teal!90!white}30.5 & \cellcolor{teal!0!white}26.3 & \cellcolor{teal!85!white}36.2 & \cellcolor{teal!24!white}43.3 & \cellcolor{teal!0!white}53.4 & \cellcolor{teal!0!white}46.1 & \cellcolor{teal!0!white}41.8 \\
         &  & RU & \cellcolor{teal!100!white}31.1 & \cellcolor{teal!42!white}36.7 & \cellcolor{teal!40!white}75.8 & \cellcolor{teal!100!white}48.3 & \cellcolor{teal!33!white}38.3 & \cellcolor{teal!55!white}37.1 & \cellcolor{teal!66!white}35.3 & \cellcolor{teal!59!white}56.3 & \cellcolor{teal!60!white}70.2 & \cellcolor{teal!92!white}59.6 & \cellcolor{teal!0!white}68.4 & \cellcolor{teal!80!white}62.6 & \cellcolor{teal!58!white}47.6 & \cellcolor{teal!95!white}46.3 & \cellcolor{teal!100!white}27.4 & \cellcolor{teal!62!white}38.3 & \cellcolor{teal!76!white}58.8 & \cellcolor{teal!35!white}43.7 & \cellcolor{teal!82!white}30.6 & \cellcolor{teal!90!white}26.0 & \cellcolor{teal!0!white}36.3 & \cellcolor{teal!16!white}43.3 & \cellcolor{teal!0!white}53.4 & \cellcolor{teal!100!white}46.0 & \cellcolor{teal!24!white}41.7 \\
         &  & SW & \cellcolor{teal!50!white}31.3 & \cellcolor{teal!14!white}36.9 & \cellcolor{teal!0!white}75.9 & \cellcolor{teal!50!white}48.5 & \cellcolor{teal!66!white}38.1 & \cellcolor{teal!11!white}37.5 & \cellcolor{teal!33!white}35.5 & \cellcolor{teal!39!white}56.6 & \cellcolor{teal!100!white}70.0 & \cellcolor{teal!0!white}61.0 & \cellcolor{teal!6!white}68.3 & \cellcolor{teal!40!white}63.0 & \cellcolor{teal!74!white}47.3 & \cellcolor{teal!80!white}46.7 & \cellcolor{teal!15!white}27.6 & \cellcolor{teal!8!white}38.6 & \cellcolor{teal!0!white}59.4 & \cellcolor{teal!0!white}44.1 & \cellcolor{teal!0!white}31.0 & \cellcolor{teal!31!white}26.2 & \cellcolor{teal!0!white}36.3 & \cellcolor{teal!0!white}43.3 & \cellcolor{teal!75!white}53.3 & \cellcolor{teal!33!white}46.1 & \cellcolor{teal!56!white}41.7 \\
         &  & ZH & \cellcolor{teal!0!white}31.5 & \cellcolor{teal!0!white}37.0 & \cellcolor{teal!19!white}75.9 & \cellcolor{teal!25!white}48.5 & \cellcolor{teal!0!white}38.5 & \cellcolor{teal!0!white}37.6 & \cellcolor{teal!83!white}35.2 & \cellcolor{teal!29!white}56.7 & \cellcolor{teal!79!white}70.1 & \cellcolor{teal!14!white}60.8 & \cellcolor{teal!19!white}68.1 & \cellcolor{teal!0!white}63.4 & \cellcolor{teal!100!white}47.0 & \cellcolor{teal!57!white}47.2 & \cellcolor{teal!9!white}27.6 & \cellcolor{teal!54!white}38.4 & \cellcolor{teal!21!white}59.2 & \cellcolor{teal!30!white}43.7 & \cellcolor{teal!46!white}30.7 & \cellcolor{teal!19!white}26.2 & \cellcolor{teal!57!white}36.2 & \cellcolor{teal!41!white}43.3 & \cellcolor{teal!49!white}53.3 & \cellcolor{teal!66!white}46.1 & \cellcolor{teal!93!white}41.6 \\
        \cline{2-28}
        \hline
        \end{tabular}
    }
\end{table*}

\FloatBarrier
\section{Supplementary Analysis}\label{app:supplementary_analysis}

\subsection{Further Language-Subspace Results}\label{app:further_language_subspace_results}

\begin{figure}[h]
    \centering
    \includegraphics[width=\linewidth]{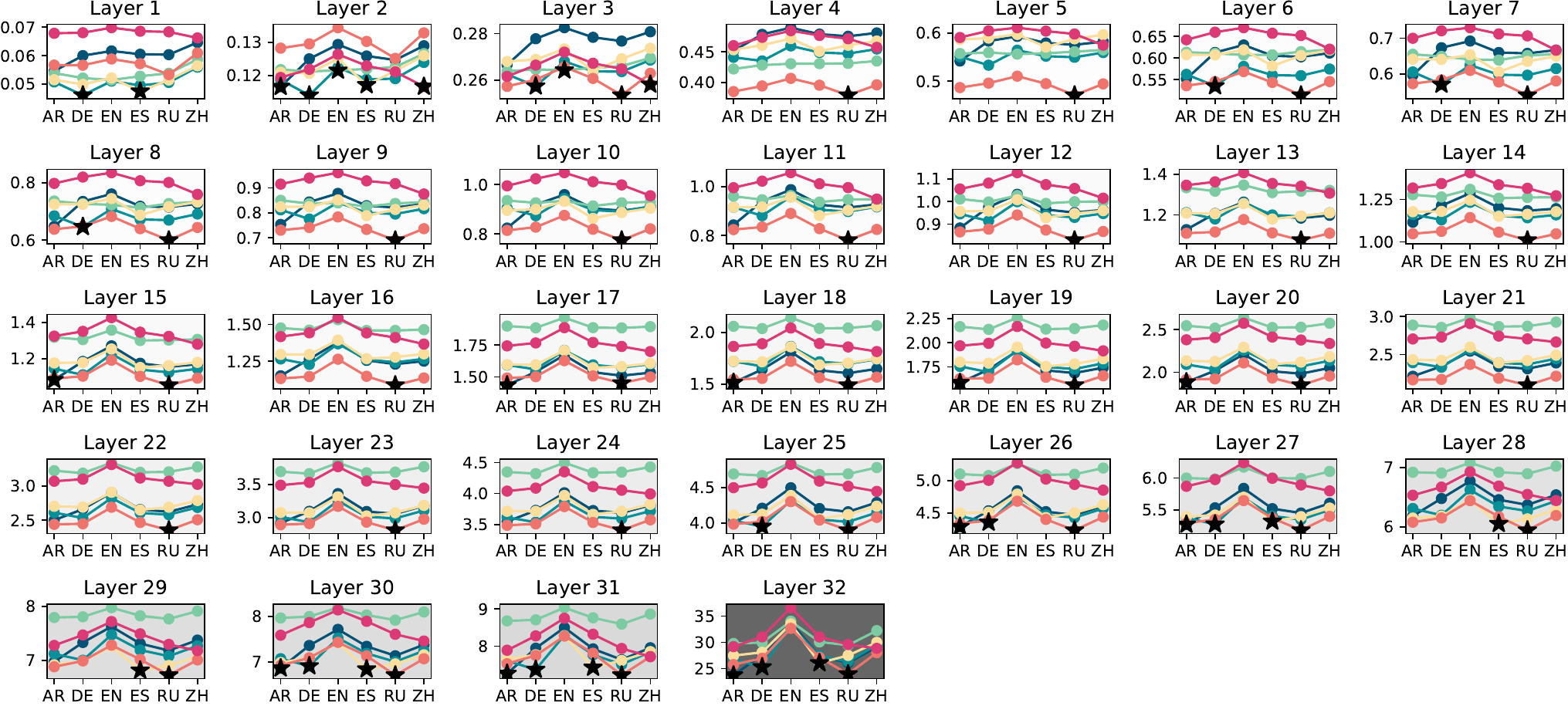}
    \caption{Language-wise mean magnitude of differences between the prompt-wise and layer-wise averaged \textbf{language-agnostic} features extracted with LSAR for a full-sized and 50\% \textbf{SparseGPT}-pruned and \textbf{mC4-calibrated} \textbf{Llama-3 8b} model. Both, the LSAR projection matrix and the feature differences, were computed over 900 prompts from the \textbf{Belebele} dataset for the six calibration/test languages. The evaluation languages are shown on the x-axis, the calibration languages are color-coded (\textcolor{ar}{AR}, \textcolor{de}{DE}, \textcolor{en}{EN}, \textcolor{es}{ES}, \textcolor{ru}{RU}, \textcolor{sw}{ZH}). The background color indicates the magnitude of the maximum deviation. A star marks the case where using the same language for calibration and evaluation results in the smallest difference after pruning.}
    \label{app:figure_lsar_a_mc4_belebele}
\end{figure}

\begin{figure}[h]
    \centering
    \includegraphics[width=\linewidth]{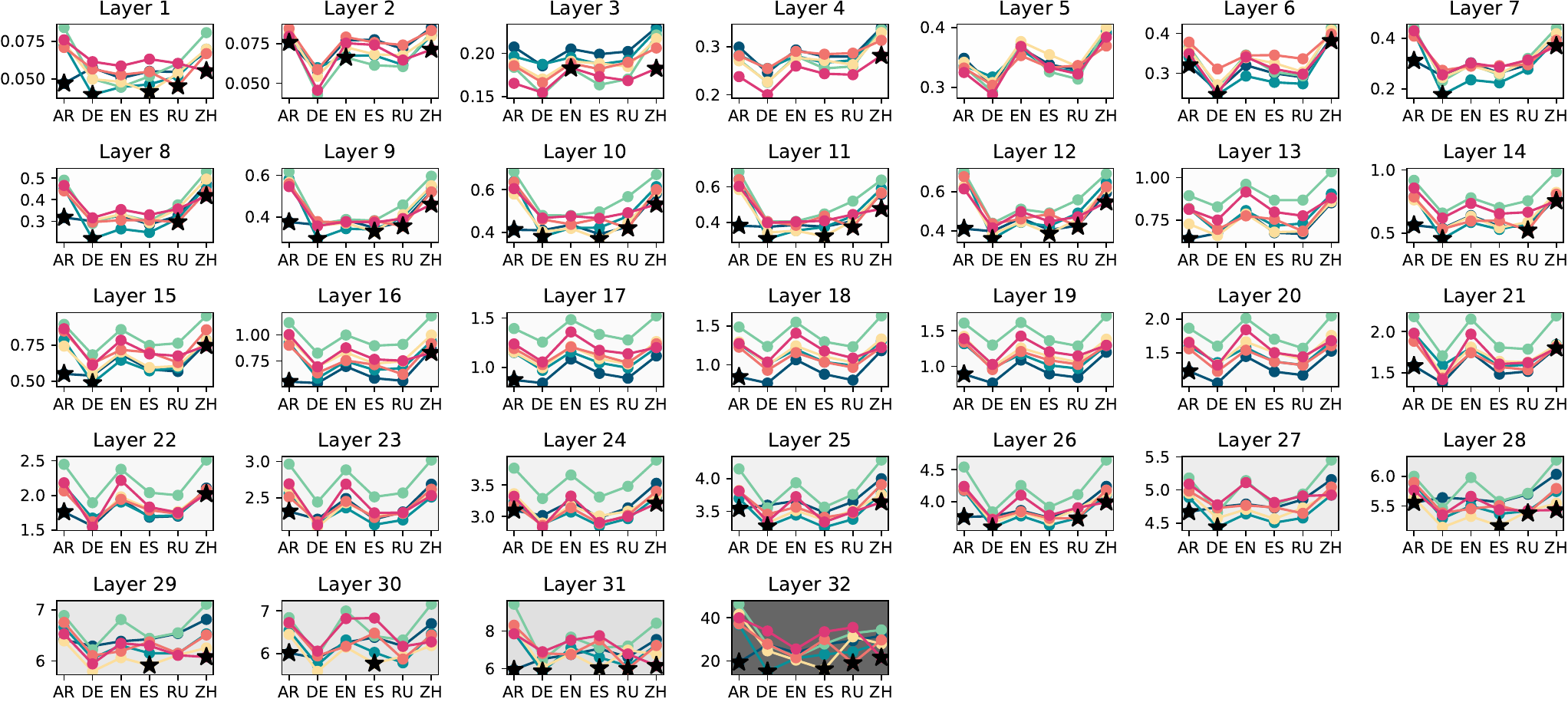}
    \caption{Language-wise mean magnitude of differences between the prompt-wise and layer-wise averaged \textbf{language-specific} features extracted with LSAR for a full-sized and 50\% \textbf{SparseGPT}-pruned and \textbf{mC4-calibrated} \textbf{Llama-3 8b} model. Both, the LSAR projection matrix and the feature differences, were computed over 900 prompts from the \textbf{Belebele} dataset for the six calibration/test languages. The evaluation languages are shown on the x-axis, the calibration languages are color-coded (\textcolor{ar}{AR}, \textcolor{de}{DE}, \textcolor{en}{EN}, \textcolor{es}{ES}, \textcolor{ru}{RU}, \textcolor{sw}{ZH}). The background color indicates the magnitude of the maximum deviation. A star marks the case where using the same language for calibration and evaluation results in the smallest difference after pruning.}
    \label{app:figure_lsar_s_mc4_belebele}
\end{figure}

\FloatBarrier
\subsection{Further Pruning Mask Similarity Results}\label{app:pms}

\begin{figure*}[h!]
    \begin{subfigure}[b]{\textwidth}
        \centering
        \includegraphics[width=5.9in]{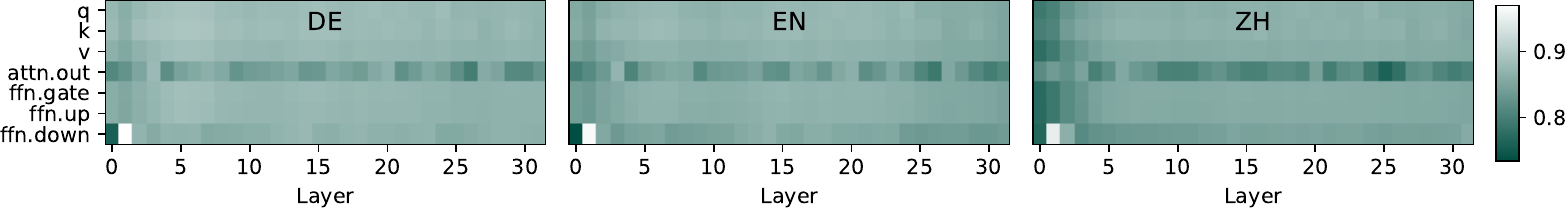}
        \caption{Intra calibration language IOUs}\label{app:fig_pms_aya-23-8b_intra}
    \end{subfigure}
    
    \begin{subfigure}[b]{\textwidth}
        \centering
        \includegraphics[width=5.9in]{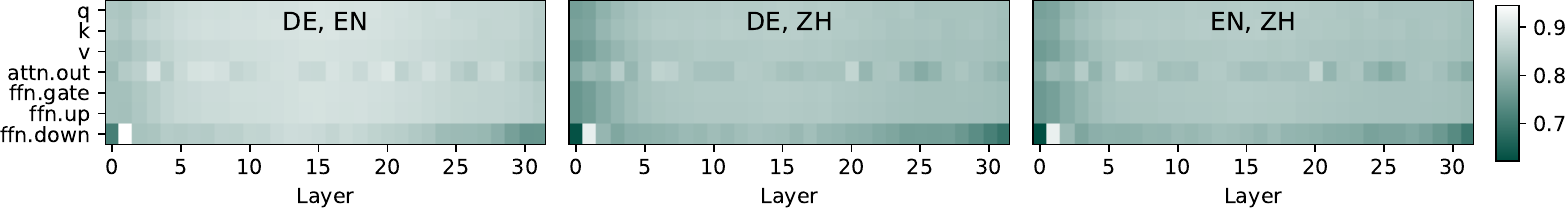}
        \caption{Inter calibration language IOUs}\label{app:fig_pms_aya-23-8b_inter}
    \end{subfigure}
    \caption{Pruning mask similarities (IoU) between EN, DE, and ZH calibrated on \textbf{mC4} and 50\% sparsity \textbf{SparseGPT}-pruned \textbf{Aya-23 8B} models. \ref{app:fig_pms_aya-23-8b_intra} IoU of pruning masks for three calibration sets of the same language. \ref{app:fig_pms_aya-23-8b_inter} IoU between pruning masks for different calibration languages. The higher IoU (indicated as a lighter color), the more similar pruning masks between different calibration languages.}
    \label{app:fig_pms_aya-23-8b}
\end{figure*}

\FloatBarrier
\subsection{Further Language Entropy Results}\label{app:lape}

\begin{figure*}[h]
    \centering
    \begin{subfigure}[b]{0.49\textwidth}
        \includegraphics[width=2.8in, left]{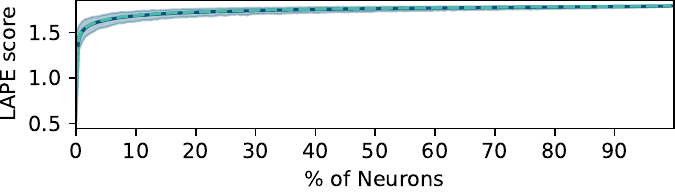}
        \vskip -0.05in
        \caption{LAPE score for full-sized and 50\% pruned model}\label{app:lape_base_aya_23_8b}
    \end{subfigure}\hfill
    \begin{subfigure}[b]{0.49\textwidth}
        \centering
        \includegraphics[width=2.8in, right]{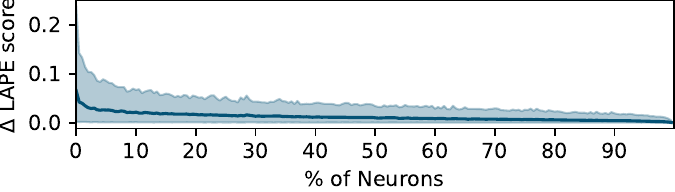}
        \vskip -0.05in
        \caption{Difference of pruned to full-sized model LAPE score}
    \end{subfigure}
    \vskip 0.1in
    \begin{subfigure}[b]{0.49\textwidth}
        \includegraphics[width=2.8in, left]{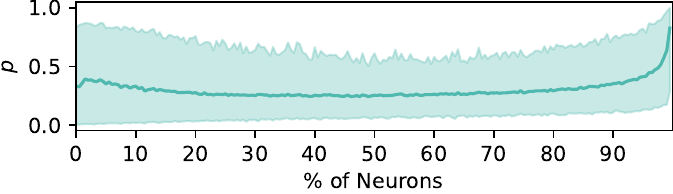}
        \vskip -0.05in
        \caption{Activation probabilities for full-sized model and DE prompts}
    \end{subfigure}\hfill
    \begin{subfigure}[b]{0.49\textwidth}
        \includegraphics[width=2.8in, right]{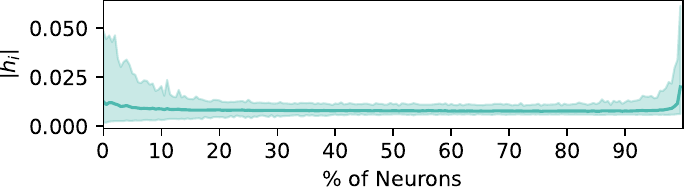}
        \vskip -0.05in
        \caption{Mean activation magnitude for full-sized model and DE prompts}
    \end{subfigure}
    \caption{Statistics for FFN neurons of a \textcolor{lape_fs}{\textbf{full-sized Aya-23 8B model}} and its \textcolor{lape_pruned}{\textbf{50\% SparseGPT-pruned}} version calibrated for DE on \textbf{mC4}. 
    Neurons are ordered by ascending LAPE score of the full-sized model as shown in \ref{app:lape_base_aya_23_8b}. 
    Additionally, LAPE score and activation probabilities get correlated by removing all neurons with an activation probability in DE that is less then the average activation probability among all languages. 
    \textit{The lower the LAPE score, the more specialized the neuron is for a particular language.}
    }
    \label{app:fig_lape_aya}
\end{figure*}

\begin{figure*}[!htbp]
    \centering
    \begin{subfigure}[b]{0.49\textwidth}
        \includegraphics[width=3in, left]{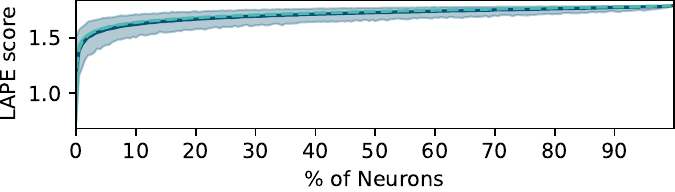}
        \vskip -0.05in
        \caption{LAPE score for full-sized and 50\% pruned model}\label{app:lape_base_llama3_8b}
    \end{subfigure}\hfill
    \begin{subfigure}[b]{0.49\textwidth}
        \centering
        \includegraphics[width=3in, right]{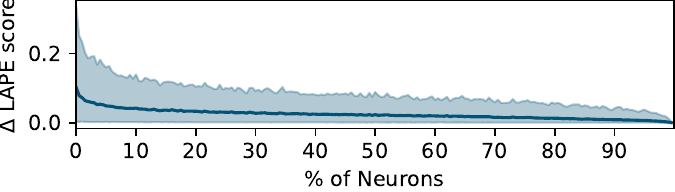}
        \vskip -0.05in
        \caption{Difference of pruned to full-sized model LAPE score}
    \end{subfigure}
    \caption{Statistics for FFN neurons of a \textcolor{lape_fs}{\textbf{full-sized Llama 3 8B model}} and its \textcolor{lape_pruned}{\textbf{50\% Wanda-pruned}} version calibrated for DE on \textbf{mC4}. 
    Neurons are ordered by ascending LAPE score of the full-sized model (\ref{app:lape_base_llama3_8b}). 
    LAPE score and activation probabilities get correlated by removing all neurons with an activation probability in DE that is less then the average activation probability among all languages. 
    }
    \label{app:fig_lape_wanda}
\end{figure*}

\end{document}